\documentclass[preprint,nopreprintline,12pt]{elsarticle}

\usepackage{hyperref}         % 解决引用不跳转的问题
\usepackage[T1]{fontenc}      % 支持扩展字符集
\usepackage[utf8]{inputenc}   % 处理 UTF-8 编码

\usepackage{booktabs}         % table
\usepackage{graphicx}         % table
\usepackage{caption}          % table
\usepackage{multirow}         % table
\usepackage{bm}               % math bold
\usepackage{makecell}         % table

\usepackage{subfig}           % figure
\usepackage{soul}
\usepackage{xcolor} % 额外引入 xcolor 以支持颜色定义
\usepackage{empheq}

\sethlcolor{yellow} % 设置高亮颜色

\soulregister\cite7 % 针对\cite命令
\soulregister\citep7 % 针对\citep命令
\soulregister\citet7 % 针对\citet命令
\soulregister\ref7 % 针对\ref命令

\hypersetup{hidelinks}      % 解决引用会出现绿色框框的问题
\PassOptionsToPackage{numbers,sort,compress}{natbib}
\makeatletter
\renewcommand{\p@subfigure}{\thefigure}  % 取消 "Figure" 前缀
\makeatletter
\usepackage{tensor}

\usepackage[ruled, vlined, linesnumbered]{algorithm2e}
\let\oldnl\nl
\usepackage{newtxtext} % 使用 Times New Roman 字体

\SetKwProg{For}{for}{ \textbf{do}}{end}
\SetKwProg{While}{while}{ \textbf{do}}{end}
\SetKwProg{If}{if}{ \textbf{then}}{end}

\usepackage{amssymb}
\usepackage{amsmath}
\usepackage{cleveref} % 引用多个公式时使用 \cref{}#
\begin{document}

\renewcommand{\figurename}{Fig.}   % 图片前面用 Fig.

\begin{frontmatter}



\title{Prior Policy Guided Dual-Agent Coordinated Manipulation Planning of Spacecraft-Manipulator System}


\author[1]{Yuhui Hu} %% Author name
\ead{huyuhui@hit.edu.cn}
\author[1]{Dong Zhou\corref{cor}}  %% Author name
\ead{dongzhou@hit.edu.cn}
\author[1]{Kaihong Ouyang}  %% Author name
\ead{25s104218@stu.hit.edu.cn}
\author[2]{Zhongliang Yu}  %% Author name
\ead{zlyu@cqu.edu.cn}
\author[3]{Jianfeng Lv}  %% Author name
\ead{ljf@lzu.edu.cn}
\author[1]{Xiangyu Shao}  %% Author name
\ead{xiangyushao@hit.edu.cn}
\affiliation[1]{organization={School of Astronautics}, %Department and Organization
            addressline={Harbin Institute of Technology}, 
            city={Harbin},
            postcode={150001}, 
          %   state={Harbin},
            country={China}}
\affiliation[2]{organization={School of Automation}, %Department and Organization
            addressline={Chongqing University}, 
            city={Chongqing},
            postcode={400044}, 
          %   state={Chongqing},
            country={China}}
\affiliation[3]{organization={School of Information Science and Engineering}, %Department and Organization
            addressline={Lanzhou University}, 
            city={Lanzhou},
            postcode={730000}, 
          %   state={Chongqing},
            country={China}}

\cortext[cor]{Corresponding author.}

%% Abstract
\begin{abstract}
The strong dynamic coupling between the manipulator and the base poses a significant challenge to maintaining spacecraft attitude stability, potentially compromising mission safety. In this paper, we propose a Dual-Agent Coordinated Manipulation Planning (DACMP) framework that simultaneously achieves high-precision end-effector pose reaching for a 6-DoF space manipulator and attitude stabilization of the base spacecraft.~To enhance learning efficiency, we present a prior policy-guided Deep Reinforcement Learning algorithm incorporating the Timestep-level Expert Switching Guidance (TESG) mechanism, thereby promoting global convergence and improving task success rates. Extensive experiments demonstrate that DACMP significantly outperforms baseline DRL algorithms in terms of task success rate and control precision.~Furthermore, the robustness of DACMP is validated under various challenging scenarios, including system constraints, environmental disturbances, and perception uncertainties. The code and simulation configurations are available on GitHub: https://github.com/HIT-YuhuiHu/DACMP.
\end{abstract}


% \begin{highlights}
% \item Dual-agent framework for spacecraft-manipulator coordinated motion planning.

% \item TESG mechanism enhances DRL training efficiency and manipulation performance.

% \item Comprehensive experiments prove DACMP's superior performance and robustness.
% \end{highlights}

%% Keywords

\end{frontmatter}



\section{Introduction}
\label{sec1}
With the increasing complexity and demands of space missions, space robots equipped with manipulators have become increasingly important due to their inherent flexibility and operability \cite{jahanshahi2024review}. Such robots are widely used in various space tasks, including target capture \cite{lei2023image}, debris removal \cite{ali2024development}, on-orbit assembly \cite{li2024optimization}, and maintenance \cite{santos2022machine}.

As the core execution module for space missions, trajectory planning aimed at precise end-effector reachability has become a primary research focus.~Traditional manipulator planning strategies are broadly classified into model-based and optimization-based approaches \cite{fallahiarezoodar2025review}.~Model-based methods, exemplified by the Generalized Jacobian Matrix (GJM) \cite{umetani1989resolved}, explicitly account for dynamic coupling to maintain end-effector accuracy in free-floating space robot systems. Recent analytical studies have also continued to advance the rigorous mathematical modeling required for complex on-orbit dynamics \cite{zhou2025saturated}. However, their heavy reliance on time-varying dynamic parameters leads to high computational complexity, hindering their application in real-time planning. Conversely, optimization-based methods, such as Differential Evolution (DE) \cite{wang2018optimal}, Constrained Particle Swarm Optimization (CPSO) \cite{wang2018coordinated}, and Rapidly-exploring Random Tree Star (RRT*) \cite{liu2025motion}, treat base reaction motion as external disturbances or kinematic constraints.~Although reducing model dependency, these algorithms are highly sensitive to parameter initialization, often leading to sub-optimal performance in dynamic space tasks.

More recently, Deep Reinforcement Learning (DRL) has emerged as a promising model-free alternative.~By interacting directly with the environment, DRL agents alleviate the dependence on accurate dynamic models and real-world data scarcity.~Algorithms such as Deep Deterministic Policy Gradient (DDPG) \cite{lillicrap2015continuous} have been successfully applied to trajectory planning tasks for space manipulators due to their capability in handling complex state spaces \cite{hu2018mrddpg, du2019learning}. Subsequently, researchers have further extended DRL-based approaches to higher-DoF space manipulators \cite{li2022constrained, zhang2024deep}, multi-manipulator systems \cite{wang2022learning, zhao2024spaceoctopus}, and increasingly complex mission scenarios \cite{blaise2023space, wei2025trajectory}. Furthermore, the latest advancements have demonstrated the versatility of DRL in handling highly dynamic trajectory planning tasks \cite{zhuang2024optimal}, as well as extending these capabilities to complex autonomous operations relying on direct visual feedback \cite{zhuang2025off, zhuang2025event}. These works demonstrate the potential of DRL to learn robust end-effector trajectory planning policies under free-floating conditions.

However, focusing exclusively on the manipulator neglects the stabilization of the base spacecraft, typically relying on idealized assumptions regarding the base dynamics.~Specifically, the fixed-base assumption implies that the spacecraft possesses sufficient control authority to perfectly counteract reaction forces \cite{d2024redundant}. In practice, this imposes excessive demands on the Attitude Control System (ACS), potentially leading to rapid fuel depletion \cite{liu2022lifetime}. Conversely, the free-floating assumption leaves the base attitude largely unregulated \cite{al2024path}. This passivity allows coupling-induced disturbances to accumulate, causing significant attitude deviations.~Such errors can cause payload misalignment, leading to Field-of-View (FOV) violations \cite{ponche2023guidance} or communication interruptions, and in severe cases, jeopardizing the operational safety of the entire system \cite{shao2022fault}.

To overcome these challenges, coordinated planning strategies that consider the coupled spacecraft-manipulator system have been developed.~Building on kinematic coupling models, the Reaction Null-Space (RNS) method \cite{nenchev2002reaction} exploits manipulator redundancy to actively suppress base vibrations. Similarly, optimization-based approaches, such as A-RPM \cite{shao2022direct}, aim to minimize base disturbances while ensuring end-effector reachability. However, these methods typically employ passive coordination, where the manipulator compromises its workspace and dexterity to accommodate the base. Furthermore, the heavy dependence on accurate dynamic models and the resulting computational burden often preclude real-time application.

To achieve active coordination, researchers have employed DRL to directly control the coupled system. Srivastava et al. \cite{srivastava2023deep} incorporated the spacecraft's reaction wheels into the planning framework, employing Proximal Policy Optimization (PPO) for a 9-DoF space robot to achieve simultaneous target reaching and base stabilization.~Nevertheless, this formulation suffers from the curse of dimensionality. To improve learning efficiency, advanced exploration frameworks such as EfficientLPT \cite{cao2023reinforcement} and HybridEL \cite{hu2025deep} have been proposed specifically for manipulator planning. However, directly extending these methods to the coupled spacecraft-manipulator system remains non-trivial due to the exponentially expanded state space. Consequently, the issue of slow convergence confines existing works to simple point-to-point motions, hindering their ability to perform complex coordinated manipulation planning tasks.

\begin{figure}[!htb]
     \centering
     \includegraphics[trim=18 18 18 18, clip, ]{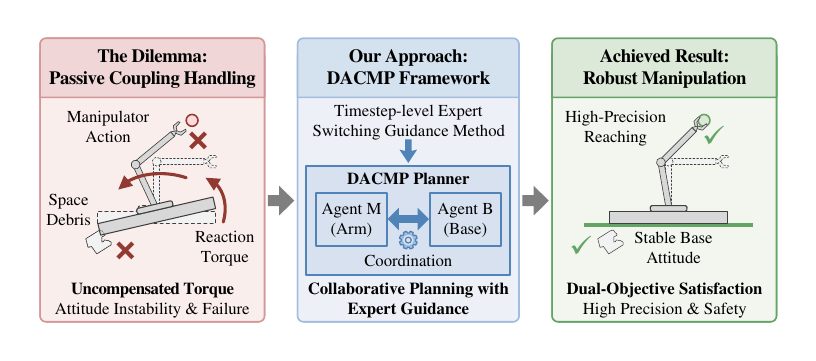}
     \caption{Conceptual illustration of the proposed Dual-Agent Coordinated Manipulation Planning (DACMP) framework.}
     \label{fig1}
\end{figure}

To solve those problems mentioned above, we propose a Dual-Agent Coordinated Manipulation Planning (DACMP) framework. This specialized architecture establishes a decoupled planning paradigm for spacecraft-manipulator systems. The schematic overview of the proposed method is illustrated in Fig.~\ref{fig1}.~The framework comprises two PPO-based reinforcement learning agents. One agent controls the 6-DoF manipulator to achieve the desired end-effector pose, while the other regulates the three-axis attitude of the base spacecraft to maintain overall system stability.~Furthermore, to improve training efficiency in high-dimensional state and action spaces, we introduce a timestep-level guidance strategy, termed Timestep-level Expert Switching Guidance (TESG). TESG exploits a non-optimal prior policy to guide learning, thereby effectively mitigating mutual interference during training and significantly improving global convergence properties.~The main contributions of this paper are summarized as follows:
\begin{itemize}
  \item We establish a decoupled planning paradigm for spacecraft-manipulator systems via the DACMP framework. This specialized dual-agent architecture effectively isolates manipulation from base stabilization, overcoming performance bottlenecks caused by strong dynamic coupling.
  \item We propose the Timestep-level Expert Switching Guidance (TESG) mechanism. By formulating a timestep-level guidance logic, TESG avoids residual action compensation and mitigates mutual interference during dual-agent exploration, significantly boosting learning efficiency.
  \item We demonstrate that the proposed approach yields significant performance improvements. Extensive experiments validate that DACMP achieves superior convergence, much higher success rates, and enhanced accuracy compared to baselines, while exhibiting strong robustness against system disturbances.
\end{itemize}

The rest of this article is organized as follows. Section \ref{sec2} presents the model of the coupled spacecraft-manipulator system and the formulation of the Dual-Agent Markov Decision Process (MDP). Section \ref{sec3} introduces the proposed approach, including the DACMP framework and the design of the prior policy guided mechanism. Section \ref{sec4} demonstrates the effectiveness and robustness of the DACMP framework through extensive experiments.~Section \ref{sec5} concludes this article.

\section{Problem Statement}
\label{sec2}
\subsection{Kinematic and dynamic modeling of coupled spacecraft-manipulator system}
\label{subsec21}

The coupled spacecraft-manipulator system consists of a base spacecraft with attitude control capabilities and a space manipulator, as shown in Fig.~\ref{fig2}.~The attitude of the base spacecraft is adjusted via three-axis control torques.~The dynamic model of this coupled system can be formulated as follows:
\begin{figure}[!htb]
     \centering
     \includegraphics[trim=18 18 18 20, clip, ]{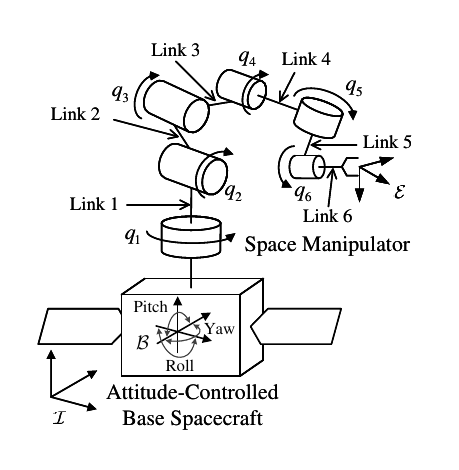}
     \caption{Model of the spacecraft-manipulator system.}
     \label{fig2}
\end{figure}

\begin{equation}
  \label{eq1}
  \left[
    \begin{matrix}
      M_\text{bb} & M_\text{bm} \\
      M_\text{mb} & M_\text{mm}
    \end{matrix}
  \right]
  \left[
    \begin{matrix}
      \dot{\omega_\text{b}} \\
      \ddot{q}
    \end{matrix}
  \right] + 
  \left[
    \begin{matrix}
      C_\text{b} \\
      C_\text{m}
    \end{matrix}
  \right]
  =
  \left[
    \begin{matrix}
      \tau_\text{b} \\
      \tau_\text{m}
    \end{matrix}
  \right]
\end{equation}
where $\omega_\text{b} \in \mathbb{R}^3$ denotes the angular velocity of the base spacecraft, and $q \in \mathbb{R}^6$ represents the joint angles of the manipulator. $M_\text{bb}$ and $M_\text{mm}$ are the equivalent inertia matrix of the base and the inertia matrix of the manipulator, respectively, while $M_\text{bm}$ and $M_\text{mb}$ represent the coupled inertia matrices between the base and the manipulator. $C_\text{b}$ and $C_\text{m}$ denote the Coriolis and centrifugal terms. Furthermore, $\tau_\text{b}$ and $\tau_\text{m}$ represent the control torques applied to the base spacecraft and the manipulator joints, respectively. By decomposing Eq.~\eqref{eq1}, the rotational dynamics of the base spacecraft can be explicitly expressed as:
\begin{equation}
  M_\text{bb}\dot{\omega_\text{b}} = \tau_\text{b} - M_\text{bm}\ddot{q} - C_\text{b}
\end{equation}
this equation highlights the dynamic coupling, where the term $M_\text{bm}\ddot{q}$ acts as a reaction torque disturbance on the base, necessitating robust manipulation planning for stability.

To formulate the kinematic model, we introduce three coordinate frames: the inertial frame $\mathcal{I}$, the base frame $\mathcal{B}$, and the end-effector frame $\mathcal{E}$. The pose of the end-effector relative to the inertial frame can be described as:
\begin{equation}
  ^{\mathcal I}\!T_{\mathcal E} = \,^{\mathcal I}\!T_{\mathcal B}({\varPhi}_\text{b})\, \,^{\mathcal B}\!T_{\mathcal E}({q})
\end{equation}
where $\varPhi_\text{b}$ denotes the attitude of the base spacecraft.~Differentiating the end-effector pose with respect to time yields the expressions for the linear and angular velocities as follows:
\begin{equation}
  \left[
    \begin{matrix}
      v_\text{e} \\
      \omega_\text{e}
    \end{matrix}
  \right]
  =
  \left[
    \begin{matrix}
      J_\text{b}\left(q\right) & J_\text{m}\left(q\right)
    \end{matrix}
  \right]
  \left[
    \begin{matrix}
      \omega_\text{b} \\
      \dot{q}
    \end{matrix}
  \right]
\end{equation}
where $J_\text{b}$ and $J_\text{m}$ represent the base and manipulator Jacobians, respectively. The composite matrix $\left[ \begin{matrix} J_\text{b}\left(q\right) & J_\text{m}\left(q\right) \end{matrix} \right]$ characterizes the combined contributions of the base rotation and manipulator motion to the end-effector state.

The derived kinematic and dynamic models reveal the inherent nonlinearity and strong coupling of the spacecraft-manipulator system.~These characteristics render real-time manipulation planning via traditional model-based methods computationally prohibitive.~To overcome these limitations, we formulate the manipulation planning problem as a Markov Decision Process (MDP) and adopt a model-free learning approach.

\subsection{Dual-Agent MDP modeling}
\label{subsec22}
To characterize the coordinated planning process between the base spacecraft and the space manipulator, we formulate the space robot manipulation planning problem as a Dual-Agent Markov Decision Process (Dual-Agent MDP). In this framework, each agent executes its policy independently while sharing the environmental state. Through cooperative learning, the agents aim to achieve the joint objectives of reaching the desired end-effector pose and maintaining base attitude stability.

The Dual-Agent MDP models the system as a tuple $\mathcal{M} = \langle S, A, P, R, \gamma \rangle$, where $S$ and $A$ denote the joint state and action spaces, defined as $S = S^\text{m} \times S^\text{b}$ and $A = A^\text{m} \times A^\text{b}$, respectively. Here, the subscripts `$\text{m}$' and `$\text{b}$' correspond to the space manipulator and the base spacecraft. $P$ is the state transition function, which is implicitly defined by the dynamics of the space robot. Given the current system state $\boldsymbol{s}_t \in S$ and the joint action $\boldsymbol{a}_t \in A$, the transition probability is expressed as:
\begin{equation}
  P := P_{\boldsymbol{s} \rightarrow \boldsymbol{s}'}^{\boldsymbol{a}} = \mathbb{P}\left[ \boldsymbol{s}_{t+1} = \boldsymbol{s}' \mid \boldsymbol{s}_t = \boldsymbol{s} , \boldsymbol{a}_t = \boldsymbol{a} \right]
\end{equation}
$R = \{R^\text{m}, R^\text{b}\}$ denotes the reward functions, which depend on the system transition from state $\boldsymbol{s}$ to $\boldsymbol{s}'$. The discount factor $\gamma \in \left[0, 1\right]$ is utilized to balance immediate rewards with long-term returns to facilitate optimal policy learning.

For the coordinated spacecraft-manipulator system, distinct agents are assigned to control the manipulator and the base spacecraft, respectively. Within the Dual-Agent MDP framework, at each timestep $t$, the system observes the state $\boldsymbol{s}_t$.~The dual agents select and execute a joint action $\boldsymbol{a}_t = \left[a_{t}^\text{m}, a_{t}^\text{b}\right]$, where the actions are generated by their corresponding policies $\pi_\text{m}$ and $\pi_\text{b}$, respectively. The system then transitions to the next state $\boldsymbol{s}_{t+1}$ governed by the state transition probability $P$. Simultaneously, the manipulator and base agent receive their individual rewards $r_{t}^\text{m}$ and $r_{t}^\text{b}$, respectively. The objective of the dual-agent DRL is to determine the optimal policy $\boldsymbol{\pi}^* = \{ \pi_\text{m}^*, \pi_\text{b}^* \}$ by maximizing the expected cumulative return to each agent individually:
\begin{equation}
  \pi_i^* = \mathop{\arg \max}_{\pi_i} \mathbb{E}\left[ \sum_{t} \gamma^t r_{t}^i \right], \quad \text{for } i \in \{\text{m}, \text{b}\}
\end{equation}

\section{Method}
\label{sec3}
\subsection{Framework of Dual-Agent Coordinated Manipulation Planning}
\label{subsec31}
\begin{figure}[!htb]
     \centering
     \includegraphics[trim=24 18 24 20, clip, width=\textwidth]{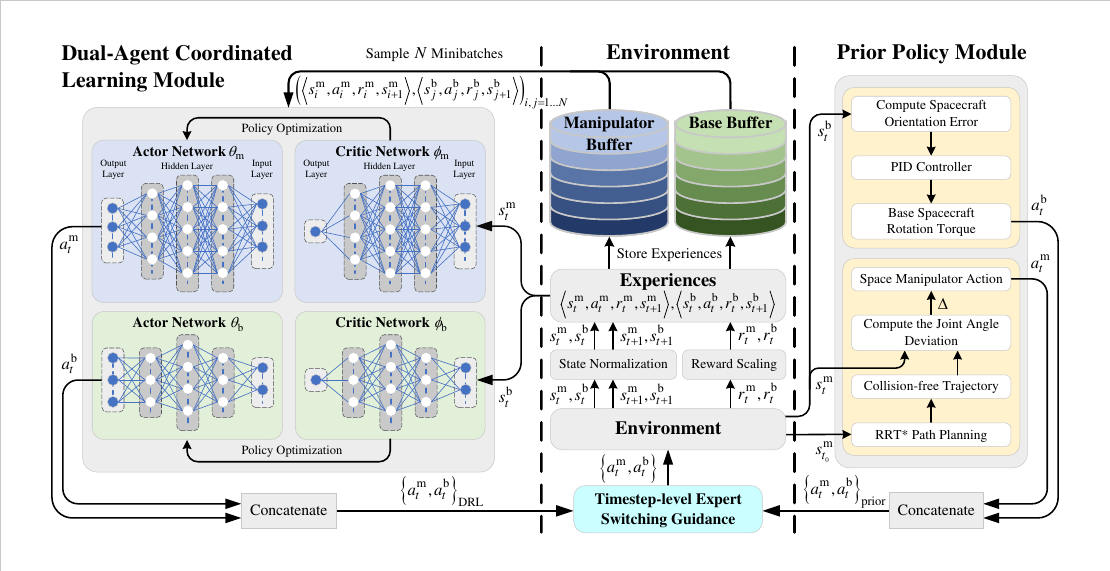}
     \caption{The overview of the proposed Dual-Agent Coordinated Manipulation Planning (DACMP) framework.}
     \label{fig3}
\end{figure}

To enhance the task adaptability of space robots while mitigating the challenges of low training efficiency and convergence uncertainty inherent in reinforcement learning, this paper proposes a novel DACMP framework. By integrating prior knowledge from classical planning and control methods with PPO-based reinforcement learning, DACMP leverages a multi-agent architecture to achieve coordinated optimization of end-effector pose reaching and base attitude stabilization.~The architecture of the proposed method is illustrated in Fig. \ref{fig3}.
% includegraphics: [trim=0 9 0 0, clip,] 左 下 右 上

\subsection{Dual-Agent reinforcement learning formulation}
\label{subsec32}
\subsubsection{Dual-Agent DRL algorithm}
\label{subsubsec321}
Because of the dynamic coupling inherent in spacecraft-manipulator systems, a single-agent approach often struggles to simultaneously ensure high-precision end-effector pose reaching and base attitude stability.~To facilitate coordinated manipulation planning, we develop a Dual-Agent DRL algorithm that governs the manipulator and the base spacecraft, respectively.~In this framework, both the manipulator agent and the base agent are immersed within a shared environment. They synchronously perceive the current environmental state and perform forward inference to generate action commands, while subsequently receiving individual rewards to independently update their policies. This decentralized architecture inherently decouples the control objectives, effectively mitigating the multi-objective interference often encountered in single-agent approaches. By assigning independent tasks to specific agents, our method reduces the dependency on complex reward scalarization techniques.

To ensure learning efficiency and stability, the proposed architecture employs a PPO-based policy gradient method.~Each agent maintains independent Actor and Critic networks to interact with the environment synchronously, with the optimization objective adopting the clipped PPO objective function:
\begin{equation}
  L_{i,t}^\text{CLIP}\left(\theta_i\right)=\mathbb{E}_t\left[\min \left(
    \rho_{t}^{i}\left(\theta_i\right)\hat{A}_{t}^{i}, \text{clip}\left(
      \rho_{t}^{i}\left(\theta_i\right), 1 - \epsilon, 1 + \epsilon
    \right)\hat{A}_{t}^{i}
  \right)\right]
  \label{eq7}
\end{equation}
where $i \in \{\text{m}, \text{b}\}$ corresponds to the space manipulator and the base spacecraft, respectively.~$\hat{A}_{t}^{i}$ denotes the advantage estimator, and $\epsilon$ represents the clipping parameter. The term $\rho_{t}^{i}\left(\theta_i\right)$ signifies the probability ratio between the new and old policies:
\begin{equation}
  \rho_{t}^{i}=\frac{\pi_{\theta_i}\left(a_{t}^{i},s_{t}^{i}\right)}{\pi_{\theta_i}^\text{old}\left(a_{t}^{i},s_{t}^{i}\right)}
\end{equation}

Distinct from the conventional PPO algorithm, our proposed DRL approach adopts an off-policy mechanism rather than the standard on-policy setting. This modification yields two primary benefits. First, the incorporation of an experience buffer reduces the frequency of model updates, thereby effectively enhancing training efficiency. Second, the off-policy architecture facilitates the integration of the Hybrid Efficient Learning (HybridEL) framework \cite{hu2025deep}, enabling the prioritized collection of successful samples during the early training stages to further secure convergence. To ensure training stability under this setting, we select a conservative clipping parameter $\epsilon=0.1$ and increase the number of gradient update steps to $K_\text{update}=90$. This configuration balances stable updates with accelerated convergence.~Detailed hyperparameters are provided in \ref{appA}, and the pseudocode for the proposed DACMP framework is presented in \ref{appB}.

\subsubsection{State and action spaces}
\label{subsubsec322}
The construction of the state space aims to provide the agent with comprehensive environmental information essential for optimization in a concise manner. For the manipulator agent, the primary objective is precise end-effector pose reaching. To this end, we adopt the continuous pose representation ($\text{CPR} \in \mathbb{R}^9 $) \cite{hu2025deep} to encode the states of the end-effector, the target, and their relative error, denoted as $\text{CPR}_\text{ee}$, $\text{CPR}_\text{tar}$, and $\text{CPR}_\text{err}$, respectively. Furthermore, to explicitly quantify tracking performance in Cartesian space, we augment the state vector with scalar Euclidean distance $e_\text{pos}$ and the orientation deviation $e_\text{ori}$. Integrating these task-space features with the 6-DoF manipulator's joint angles $q$ and angular velocities $\dot{q}$, we formulate a 41-dimensional state space for the manipulator agent:
\begin{equation}
  s_{t}^\text{m} = \left[
    \text{CPR}_\text{ee}, \text{CPR}_\text{tar}, \text{CPR}_\text{err}, e_\text{pos}, e_\text{ori}, q, \dot{q}
  \right]
\end{equation}

For the base spacecraft, since the stabilization target aligns with the initial attitude, explicit target encoding is omitted to avoid redundancy.~Instead, we utilize the continuous rotation representation $\text{CRR}_\text{b} \in \mathbb{R}^6$, which isolates the orientation component of the $\text{CPR}$, as the core observation.~To account for dynamic disturbances while maintaining state compactness, we explicitly include the joint angles and angular velocities of the first three manipulator joints, which exert the dominant coupling influence on the base attitude. Finally, by incorporating the previous base torque $\tau_{\text{b}, t-1}$ as the actuation history, we formulate a 15-dimensional state space for the base spacecraft agent:
\begin{equation}
  s_{t}^\text{b} = \left[
    \text{CRR}_\text{b}, q_{1:3}, \dot{q}_{1:3}, \tau_{\text{b}, t-1}
  \right]
\end{equation}

The definition of the action space is formulated based on the specific requirements of the manipulation planning task. For the space manipulator, the action space is defined as the joint angular velocities $a_{t}^\text{m} = \dot{q} \in \mathbb{R}^6$.~To satisfy physical constraints and ensure operational safety, the velocity command for each joint is clipped within the range of $\pm 2$ rad/s.

For the base spacecraft, the action space consists of the three-axis control torque $a_{t}^\text{b} = \tau_\text{b} \in \mathbb{R}^3$. To align with realistic engineering constraints, we bound the control output based on the capabilities of typical actuators employed in space missions.~Specifically, we enforce a rated saturation torque of 0.1 $\text{N} \cdot \text{m}$.~This specification is representative of the control authority available to standard 100 kg-class space robots \cite{de2024high}, thereby aligning the simulation constraints with realistic engineering standards.

\subsubsection{Reward function formulation}
\label{subsubsec323}
For the manipulator agent, we follow the reward design established in \cite{hu2025deep}, constructing a composite reward function that balances pose accuracy and motion smoothness. Specifically, the reward $r^{\text{m}}$ is composed of four terms: pose error penalty $p_{\text{pose}}^\text{m}$, smoothness penalty $p_{\text{smth}}^\text{m}$, orientation alignment reward $r_{\text{aln}}^\text{m}$, and completion reward $r_{\text{done}}^\text{m}$. The formulation is given by:
\begin{equation}
r^{\text{m}} = -p_{\text{pose}}^\text{m} - p_{\text{smth}}^\text{m} + r_{\text{aln}}^\text{m} + r_{\text{done}}^\text{m}
\end{equation}
where the individual reward terms are defined as follows:
\begin{align}
  p_{\text{pose}}^\text{m} &= k_{\text{pos}}^\text{m} \cdot e_\text{pos} + k_{\text{ori}}^\text{m} \cdot e_\text{ori} \\
  p_{\text{smth}}^\text{m} &= k_{\text{smth}}^\text{m} \cdot \sum_{j=1}^{6} \max\left(
    \vert \dot{q}_{j,t} - \dot{q}_{j,t-1} \vert - \delta_{\dot{q}}, 0
    \right) \\
  r_{\text{aln}}^\text{m} &= k_{\text{aln}}^\text{m} \cdot \max \left(
    \vert e_{\text{ori}, t-1} - e_{\text{ori}, t} \vert , 0
  \right) \\
  r_{\text{done}}^\text{m} &= k_{\text{done}}^\text{m} \cdot \left[
    \max \left( \frac{\varepsilon_\text{pos} - e_\text{pos}}{\varepsilon_\text{pos}} , 0\right) + 
    \max \left( \frac{\varepsilon_\text{ori} - e_\text{ori}}{\varepsilon_\text{ori}} , 0\right)
  \right]
\end{align}
In these equations, the coefficients $k_{(\cdot)}^\text{m}$ serve as weighting factors to balance the reward components, while $\varepsilon_\text{pos}$ and $\varepsilon_\text{ori}$ define the error thresholds for position and orientation errors, respectively. $\delta_{\dot{q}}$ represents the acceleration variation tolerance. The specific values are detailed in \ref{appC}.

For the base agent, the primary optimization objective is to ensure the attitude stability of the base spacecraft. To this end, we design a reward function comprising three components. The total reward function is given by:
\begin{equation}
  r^{\text{b}} = -p_{\text{att}}^\text{b} + r_{\text{var}}^\text{b} + r_{\text{done}}^\text{b}
\end{equation}
where the specific terms are defined as follows:

\begin{enumerate}
  \item[1] \textbf{Base attitude penalty}: This term is designed to penalize deviations in the base attitude. The penalty is formulated as:
  \begin{equation}
    p_{\text{att}}^\text{b} = k_{\text{att}}^\text{b} \cdot e_\text{att}
  \end{equation}
  where $e_\text{att}$ denotes the scalar attitude error of the base spacecraft.
  \item[2] \textbf{Base attitude variation reward}: This component provides immediate feedback on the temporal evolution of the base attitude error, incentivizing error reduction while penalizing divergence.~Notably, this term explicitly accounts for not only the overall scalar attitude error but also the individual deviations along the pitch, yaw, and roll axes. It is formulated as:
  \begin{equation}
    r_{\text{var}}^\text{b} = k_{\text{var}}^\text{b} \cdot 
      \left[
        \left(
          e_{\text{att}, t-1} - e_{\text{att},t}
        \right) + 
        \left(
          \varPhi_{t-1} \ominus \varPhi_t
        \right)
      \right]
  \end{equation}
  where $\varPhi_t = \left[\phi_t \; \theta_t \; \psi_t\right]$ represents the Euler angle orientation of the base spacecraft using the Z-Y-X rotation sequence. Since the base attitude variation remains limited during manipulation planning, the gimbal lock problem is effectively avoided. The difference operator $\ominus$ is defined as the reduction in the $L_1$ norm of Euler angle vector:
  \begin{equation}
    \begin{aligned}
      \varPhi_{t-1} \ominus \varPhi_{t} 
      &\triangleq 
      \|  \varPhi_{t-1} \| _1 - \| \varPhi_{t} \| _1 \\
      &= 
      \left( |\phi_{t-1}| + |\theta_{t-1}| + |\psi_{t-1}| \right) - \left( |\phi_{t}| + |\theta_{t}| + |\psi_{t}| \right)
    \end{aligned}
  \end{equation}
  \item[3] \textbf{Base completion reward}: This term is triggered when the base attitude error falls within a specified tolerance range. Once this condition is met, a positive reward is assigned to the agent, calculated as follows:
  \begin{equation}
      r_{\text{done}}^\text{b} = k_{\text{done}}^\text{b} \cdot \max \left( \frac{\varepsilon_\text{att} - e_\text{att}}{\varepsilon_\text{att}} , 0\right)
  \end{equation}
  where $\varepsilon_\text{att}$ denotes the attitude error threshold for the base spacecraft.
\end{enumerate}

The detailed specifications of hyperparameters for both the manipulator and base agents are provided in \ref{appC}.

\subsection{Timestep-level Expert Switching Guidance method}
\label{subsec33}
\subsubsection{Model-based prior policy design}
\label{subsubsec331}
In high-dimensional task spaces, learning-based algorithms are prone to getting trapped in poor local optima during the early training stages, often resulting in premature convergence and suboptimal performance. Furthermore, the inherent dynamic coupling of the dual-agent architecture amplifies training instability. To navigate this complex optimization landscape, we integrate prior policies during early training.~This external guidance steers the agent away from sub-optimal regions, ensuring high-quality task performance. Specifically, for the trajectory planning task of the space manipulator, we employ a sampling-based global planning algorithm to generate collision-free reference trajectories.~Conversely, for the base spacecraft, a PID controller is adopted to provide reference actions for attitude stabilization. A key advantage of these methods is their ability to provide baseline manipulation planning strategies and ensure basic task feasibility without requiring precise dynamic models.

For the trajectory planning of the space manipulator, we employ the RRT* algorithm \cite{qi2025research} as the prior policy. As an asymptotically optimal sampling-based planner, RRT* improves upon the standard RRT by introducing rewiring mechanisms to continuously refine the tree structure. This capability allows the algorithm to converge to a cost-optimal path, making it an ideal candidate for providing high-quality reference trajectories to guide the DRL agent efficiently.~To reduce computational complexity, RRT* planning is performed in the joint space under a fixed-base assumption.~This simplification avoids the inefficiency of solving kinematics with a time-varying GJM while also eliminating the prohibitive cost of computing coupled base dynamics during every sampling step.~This approximation is sufficient for guiding efficient exploration without requiring high-precision execution. Consequently, the generated continuous joint trajectory is numerically differentiated to obtain angular velocities aligned with the DRL action space.

Regarding base attitude stabilization, we adopt a discrete PID controller as the prior policy due to its model-free efficiency.~The control law computes the required three-axis torque $a_{t}^\varPhi$ based on the Euler angle error $e_{t}^\varPhi$:
\begin{equation}
  a_{t}^\varPhi = K_p \cdot e_{t}^\varPhi + K_i \cdot \sum _{k=0}^{t} e_{k}^\varPhi + K_d \cdot \left(e_{t}^\varPhi - e_{t-1}^\varPhi\right)
\end{equation}
where the gains are set to $K_p=15$, $K_i=2.5$, and $K_d=800$ to ensure robust tracking. To satisfy actuator limitations, the final output torque is clipped to the range of $[-0.1, 0.1]$ $\text{N} \cdot \text{m}$.

\subsubsection{Timestep-level Expert Switching Guidance mechanism}
\label{subsubsec332}
To effectively leverage prior knowledge, conventional prior guidance methods typically adopt a linear blending strategy \cite{cao2023reinforcement}, formulated as:
\begin{equation}
  a_k\left(s\right) = w_k \cdot a_{\theta_k} \left(s\right) + \left(1-w_k\right) \cdot a_{\text{prior}}\left(s\right)
\end{equation}
where $a_k$ represents the blended policy at the $k$-th training episode, $a_{\theta_k}$ denotes the agent's learned policy, and $a_\text{prior}$ is the prior policy. Typically, the fusion coefficient $w_k \in [0, 1]$ increases monotonically with the training episode $k$. This schedule facilitates a gradual transition from prior-dominated guidance in the early stages to a purely learning-based policy as training progresses. Despite the simplicity of linear blending, it encourages the agent to learn compensatory actions to correct the residuals of the prior policy, rather than the autonomous capability for independent manipulation planning \cite{nakhaei2024residual}. Consequently, the agent fails to acquire a standalone policy for independent task execution.

To circumvent these aforementioned limitations, we propose the Timestep-level Expert Switching Guidance (TESG) method, where the prior policy serves as the expert to guide the learning process. TESG probabilistically selects an action from either the learned policy or the prior policy at each timestep. The composite policy model of TESG is formulated as:
\begin{equation}
  a_k\left(s\right) = \xi_k  \cdot a_{\theta_k} \left(s\right) + \left(1-\xi_k\right) \cdot a_{\text{prior}}\left(s\right)
\end{equation}
where $\xi_k \in \{0, 1\}$ is a Bernoulli switching variable governed by $\mathbb{P}\left(\xi_k=1\right)=p\left(k\right)$. Here, $p(k)$ denotes the selection probability of the learned policy, follows a piecewise linear increase schedule:
\begin{equation}
  p\left(k\right) =
  \begin{cases}
    0.3 + 0.5 \times \frac{k}{k_\text{g}}, & k \leq k_\text{g} \\
    1, & k > k_\text{g}
  \end{cases}
  \label{eq24}
\end{equation}
where $k_\text{g} = 15$~represents the total number of guidance epochs.~Under this schedule, the prior policy dominates early training to provide stable exploration guidance. More critically, whenever the learned policy is selected, the resulting reward is attributed solely to the independent action of the agent.~This decoupling mechanism effectively prevents the agent from learning to merely compensate for the prior policy residuals. To balance exploration and exploitation, TESG operates in two phases. During the initial $k_\text{g}$ epochs, the prior policy dominates to ensure safe, guided exploration. As $k$ increases, the linear growth of $p(k)$ enables the agent to progressively exploit its own learned policy. For $k > k_\text{g}$, the prior guidance is permanently deactivated, ensuring that coordinated motion planning relies entirely on the autonomous DRL policies to approach asymptotic optimality.

\section{Experiments}
\label{sec4}

\subsection{Experimental setup}
\label{subsec41}
\begin{figure}[t]
  \centering
  \includegraphics{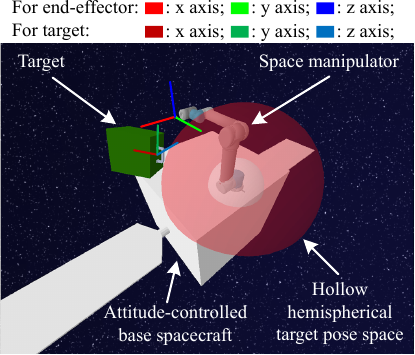}
  \caption{Simulation environment of the spacecraft-manipulator system.}
  \label{fig4}
\end{figure}

The simulation environment for space robot manipulation planning is established within the PyBullet physics engine, as illustrated in Fig.~\ref{fig4}.~The system comprises a base spacecraft and a space manipulator.~Specifically, the base is modeled as a cubic satellite equipped with bilateral solar panels, while a UR5 robotic arm serves as the manipulator.~The arm is rigidly mounted on the base, and the integrated system operates within a simulated zero-gravity environment.~The detailed physical parameters for both components are summarized in Table \ref{table1}.

% includegraphics: [trim=0 9 0 0, clip,] 左 下 右 上

\begin{table}[t]
  \caption{Parameters of the spacecraft-manipulator system.}
  \label{table1}
  \footnotesize
  \centering
  \begin{tabular}{l l l}
    \toprule
    Component & Parameters & Values \\     % Table header
    \midrule
    \multirow{5}{*}{Base Spacecraft}
    & Mass [kg] & 100 \\
    & Offset vector [m] & $(0, 0, 0)$ \\
    & Size [m] & $(1.2, 1.2, 1.2)$ \\
    & Inertia matrix [$\text{kg} \cdot \text{m}^2$] & $\text{diag}(41.6, 52.9, 52.9)$ \\
    & Control torque range [$\text{N} \cdot \text{m}$] & $(\pm 0.1, \pm 0.1, \pm 0.1)$ \\
    % \hspace*{\fill} \\
    \addlinespace
    % \midrule
    \multirow{3}{*}{Manipulator}
    & Offset vector [m] & $(0, -0.4, 0.6)$ \\
    & Joint angle range [rad] & $(\pm 2\pi, \pm 2\pi, \pm 2\pi, \pm \pi, \pm \pi, \pm \pi)$ \\
    & Joint angular velocity range [$\text{rad} / \text{s}$] & $(\pm 2, \pm 2, \pm 2, \pm 2, \pm 2, \pm 2)$ \\
    \bottomrule
  \end{tabular}
\end{table}

For the manipulation planning task configuration, target poses are randomly initialized within the dexterous workspace of the manipulator. The workspace is defined as a hollow hemisphere centered at the manipulator base, bounded by an inner radius of 0.25 m and an outer radius of 0.65 m.

Quantitative performance is evaluated over $N_\text{e} = 1000$ independent evaluation episodes. We report the Average Position Error (APE), Average Orientation Error (AOE), and Average Base Attitude Error (ABAE) as the primary metrics. To capture terminal stability, these metrics are computed by averaging the end-effector position error $e_\text{pos}$, orientation error $e_\text{ori}$, and base attitude error $e_\text{att}$ over the final 10 timesteps, denoted as the observation window $\mathcal{T}_{\text{end}}$. The explicit formulas are defined as follows:
\begin{align}
  \text{APE} = \frac{1}{N_{\text{e}}} \sum_{i=1}^{N_{\text{e}}} \left( \frac{1}{10} \sum_{t \in \mathcal{T}_{\text{end}}} e_{\text{pos}, t}^i \right) \\
  \text{AOE} = \frac{1}{N_{\text{e}}} \sum_{i=1}^{N_{\text{e}}} \left( \frac{1}{10} \sum_{t \in \mathcal{T}_{\text{end}}} e_{\text{ori}, t}^i \right) \\
  \text{ABAE} = \frac{1}{N_{\text{e}}} \sum_{i=1}^{N_{\text{e}}} \left( \frac{1}{10} \sum_{t \in \mathcal{T}_{\text{end}}} e_{\text{att}, t}^i \right)
\end{align}

Additionally, the Average Success Rate (ASR) is incorporated to assess overall reliability. A task is classified as successful only if the system simultaneously satisfies the end-effector position error $e_\text{pos} \leq \varepsilon_\text{pos}$ $(0.05 \text{ m})$, the orientation error $e_\text{ori} \leq \varepsilon_\text{ori}$ $(0.1 \text{ rad})$, and the base attitude error $e_\text{att} \leq \varepsilon_\text{att}$ $(0.05 \text{ rad})$. These conditions must be maintained for 10 consecutive timesteps, with each timestep lasting 0.1 seconds. The ASR is formally calculated using an indicator function across the $N_{\text{e}}$ episodes:
\begin{equation}
   \text{ASR} = \frac{1}{N_{\text{e}}} \sum_{i=1}^{N_{\text{e}}} \mathbb{I}(S_i)
\end{equation}
where $\mathbb{I}$ represents the indicator function, returning 1 if the success criteria $S_i$ are met for episode $i$, and 0 otherwise.

\subsection{Performance comparison}
\label{subsec42}
\begin{figure}[!htb]
  \centering
  \subfloat[Average Success Rate (ASR)]{
    \includegraphics[trim=0 0 34 12, clip, width=0.47\textwidth]{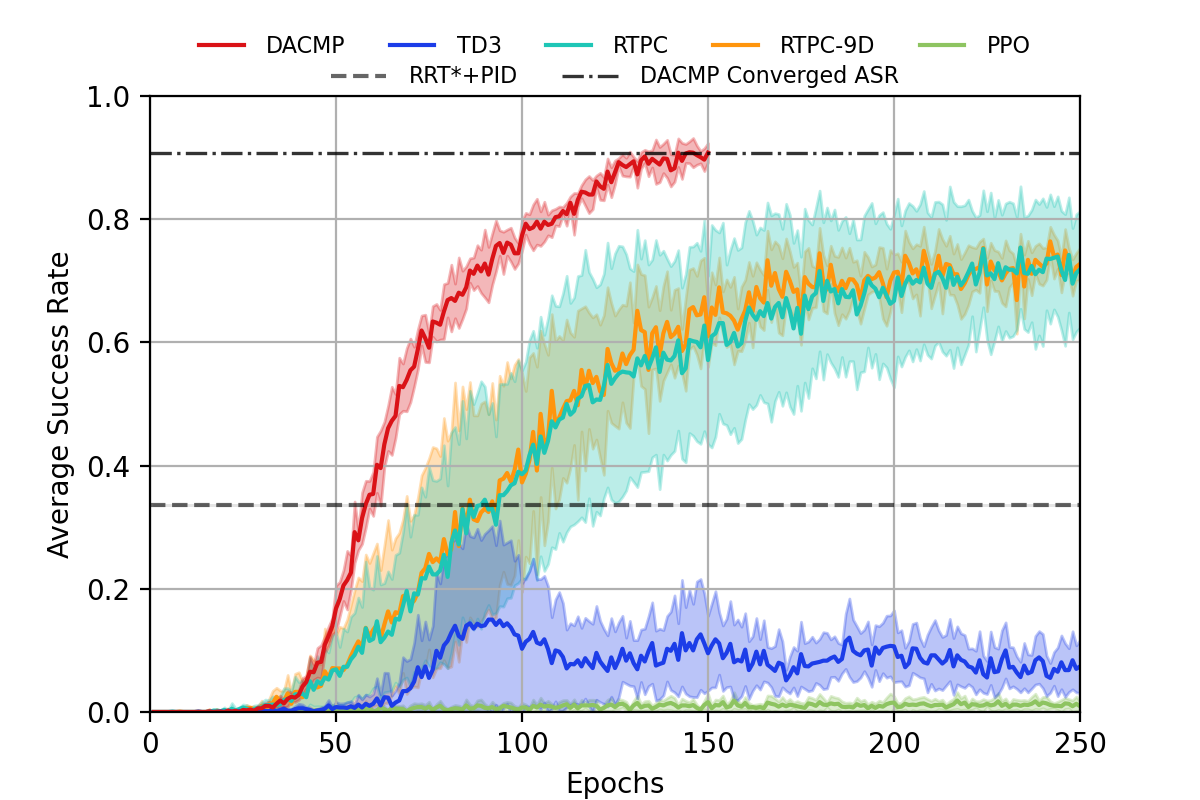}
    \label{fig5_ul}
  }
  % \hfill
  \subfloat[Average Base Attitude Error (ABAE)]{
    \includegraphics[trim=0 0 34 12, clip, width=0.47\textwidth]{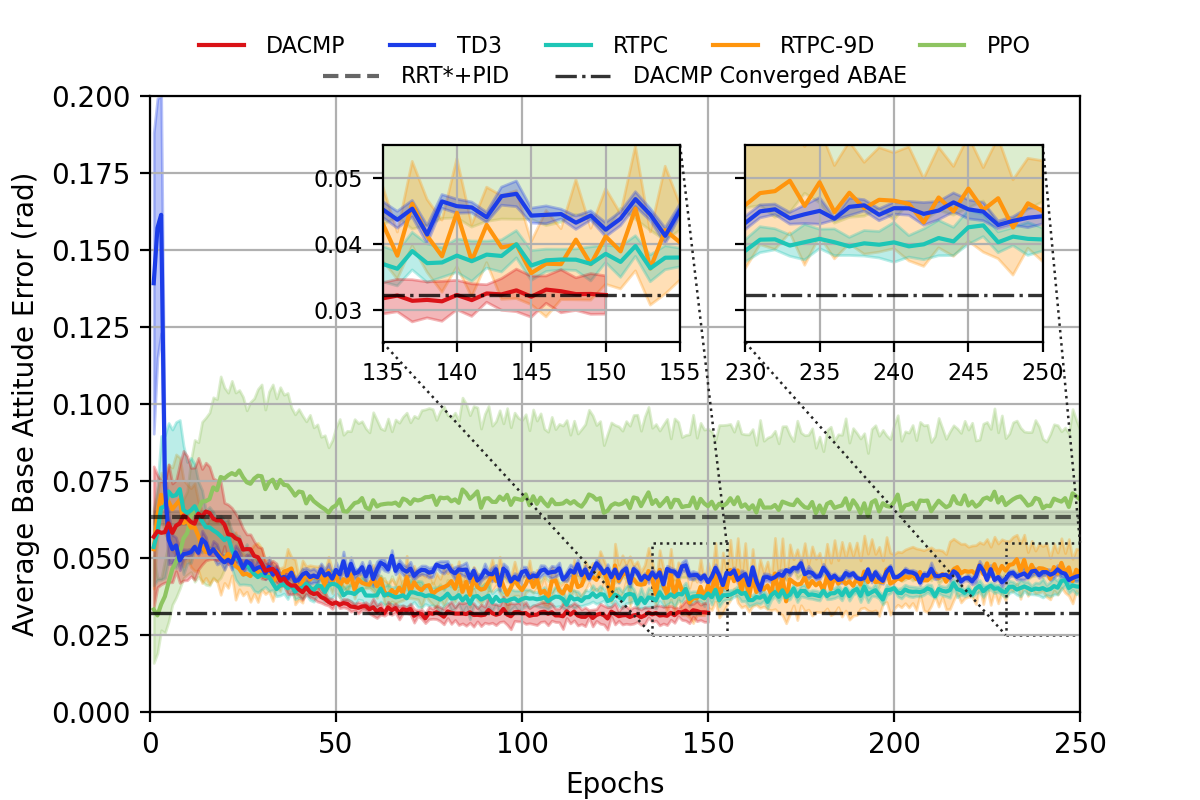}
    \label{fig5_ur}
  } \\
  \subfloat[Average Position Error (APE)]{
    \includegraphics[trim=0 0 34 12, clip, width=0.47\textwidth]{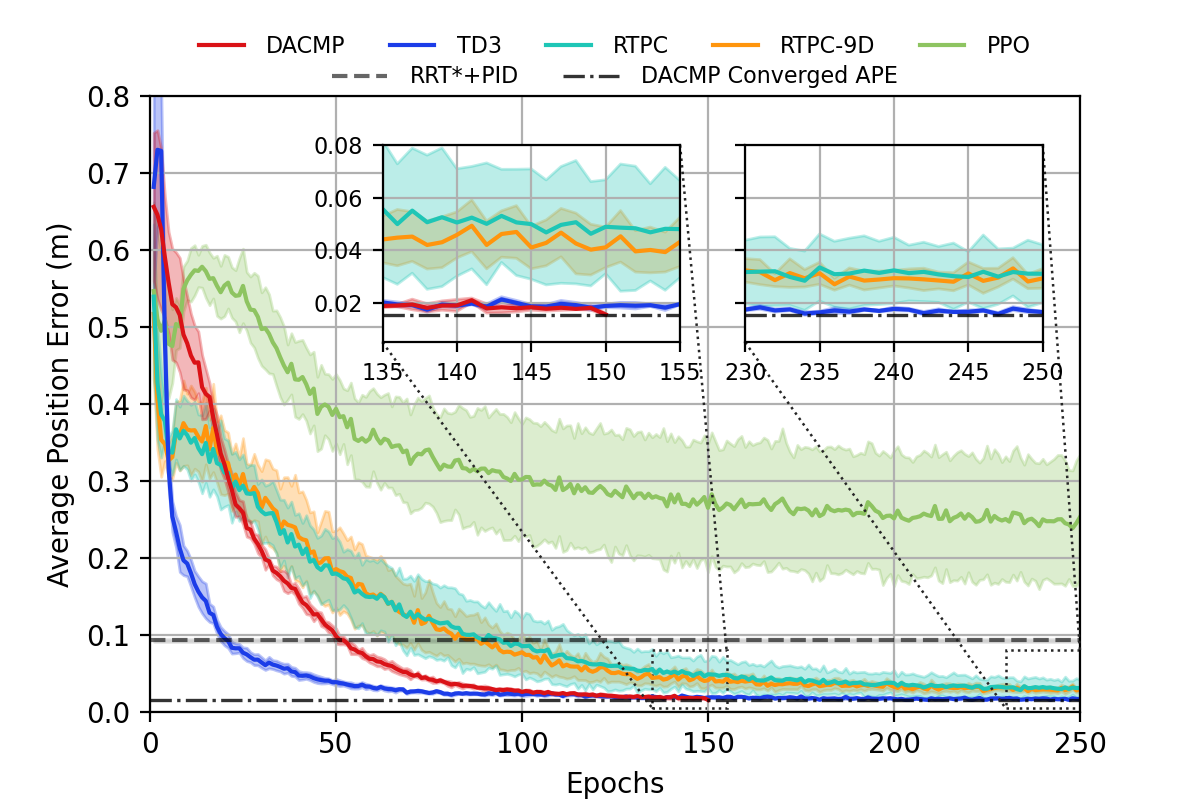}
    \label{fig5_ll}
  }
  % \hfill
  \subfloat[Average Orientation Error (AOE)]{
    \includegraphics[trim=0 0 34 12, clip, width=0.47\textwidth]{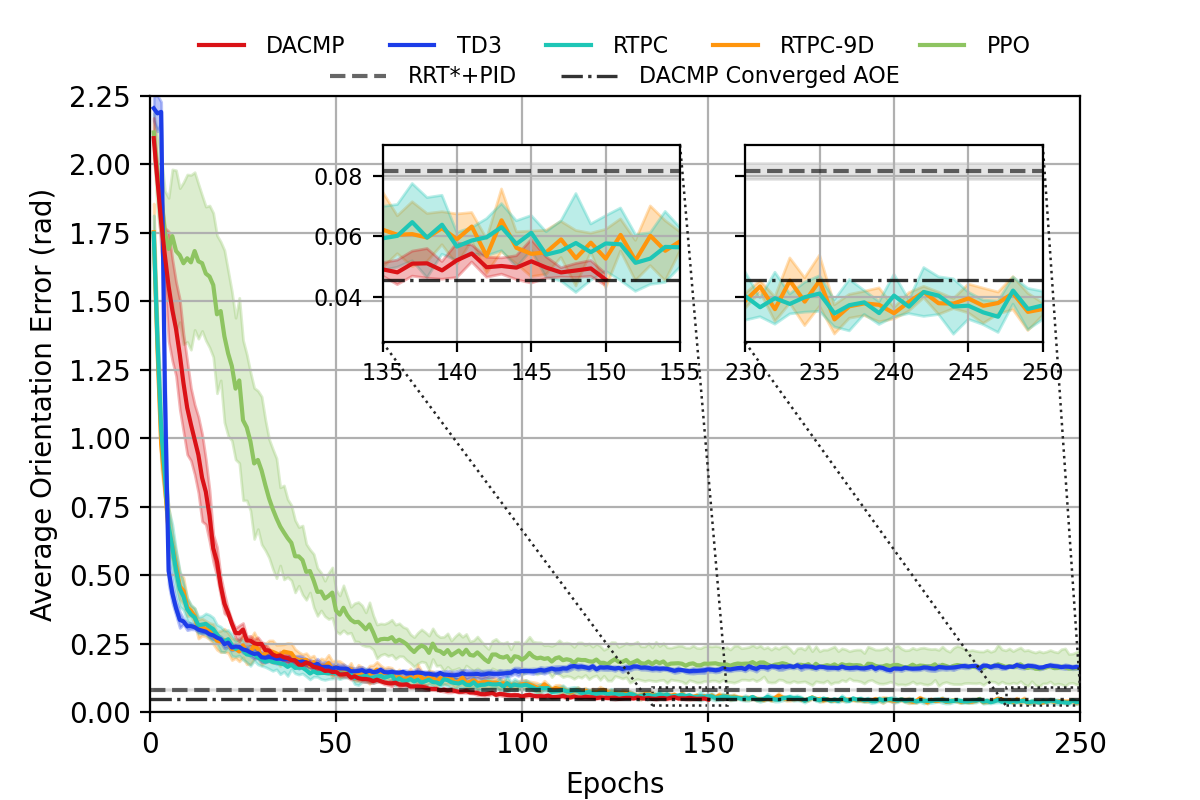}
    \label{fig5_lr}
  }
  \caption{Performance comparison of the proposed DACMP framework against baseline algorithms. The solid lines represent the mean values across 5 random seeds, while the shaded regions indicate the standard deviation. The horizontal dashed lines represent the performance of the non-learning RRT*+PID baseline, and the dash-dotted lines represent the final converged performance of the DACMP algorithm.}
  \label{fig4-s4f2}
\end{figure}

In this section, we evaluate the performance of the proposed DACMP framework against four DRL-based baselines and a purely model-based approach. We select Twin Delayed Deep Deterministic Policy Gradient (\textbf{TD3}) \cite{song2024trajectory} and \textbf{RTPC} \cite{hu2025deep} to represent off-policy and on-policy algorithms, respectively. In these setups, the DRL agents are dedicated solely to the manipulator trajectory planning task, while the base spacecraft is stabilized using the PID controller described in Section \ref{subsubsec331}. Additionally, we extend the RTPC algorithm to a 9-DoF configuration, denoted as \textbf{RTPC-9D}, which employs a single agent to simultaneously control both the manipulator and the base spacecraft.~Another single-agent \textbf{PPO}-based method \cite{srivastava2023deep} is also included for comparison.~We preserve the original action space and reward function of this method, while augmenting the state space with relative target pose information. The baseline hyperparameters are listed in Table \ref{tableap1} for reproducibility. Finally, we compare our method against a non-learning baseline, which relies on \textbf{RRT*} and \textbf{PID} without any learning component.~The performance comparison curves are depicted in Fig. \ref{fig4-s4f2}, and the quantitative results are summarized in Table \ref{table2}.

% includegraphics: [trim=0 9 0 0, clip,] 左 下 右 上

\begin{table}[b]
  \caption{The performance of different algorithms on manipulation planning.}
  \label{table2}
  \footnotesize
  \centering
  \resizebox{\textwidth}{!}{%
    \begin{tabular}{l c c c c}
      \toprule
      Algorithm   & ASR & ABAE [rad] & APE [m] & AOE [rad] \\     % Table header
      \midrule
      DACMP (ours) & $\textbf{0.9080} \pm 0.0154$ & $\textbf{0.0322} \pm 0.0029$ & $\textbf{0.0156} \pm 0.0005$ & $0.0457 \pm 0.0020$ \\
      TD3         & $0.0740 \pm 0.0428$ & $0.0442 \pm 0.0012$ & $0.0166 \pm 0.0002$ & $0.1621 \pm 0.0054$ \\
      RTPC        & $0.7187 \pm 0.0943$ & $0.0407 \pm 0.0017$ & $0.0311 \pm 0.0110$ & $0.0372 \pm 0.0046$ \\
      RTPC-9D     & $0.7273 \pm 0.0262$ & $0.0450 \pm 0.0077$ & $0.0294 \pm 0.0038$ & $\textbf{0.0360} \pm 0.0021$ \\
      PPO         & $0.0093 \pm 0.0071$ & $0.0690 \pm 0.0226$ & $0.2532 \pm 0.0812$ & $0.1636 \pm 0.0622$ \\
      RRT*+PID    & $0.3367 \pm 0.0000$ & $0.0632 \pm 0.0021$ & $0.0929 \pm 0.0019$ & $0.0814 \pm 0.0026$ \\
      \bottomrule
    \end{tabular}
  }
\end{table}

As illustrated in Fig. \ref{fig5_ul}, the proposed DACMP framework demonstrates superior convergence performance, reaching stability around Epochs 150 and achieving a final ASR of 90.8\%.~In contrast, both RTPC and RTPC-9D require approximately 250 Epochs to converge, yielding a significantly lower ASR around 73\%. Meanwhile, TD3 and PPO struggle to establish effective policies in manipulation planning tasks, with ASR stagnating below 10\%.

A detailed inspection of the error curves reveals distinct learning characteristics among the algorithms.~TD3, rather than DACMP, exhibited the most rapid reduction in APE during the initial training phase. This can be attributed to its off-policy nature, where high sample efficiency is achieved by reusing experiences from the replay buffer. Since position errors are easier to minimize, the TD3 agent tends to greedily select actions that yield an immediate reduction in position error at the expense of orientation alignment. Consequently, as shown in Fig. \ref{fig5_lr}, the AOE for TD3 deteriorated in the later stages, leading to a decrease in overall task performance.

Conversely, RTPC and RTPC-9D displayed the opposite trend, with a rapid early convergence in AOE, which appeared to hinder the optimization of APE. The error curves clearly evidence a trade-off behavior among these baselines, where proficiency in one metric is associated with significant deficits in another. Notably, the single-agent RTPC-9D failed to outperform the hybrid architecture (RTPC with PID) for base attitude precision.~This confirms that a single agent struggles to manage coupled dynamics in complex planning tasks.

In summary, the evaluated baselines suffer from distinct architectural bottlenecks. TD3 exhibits a short-sighted exploitation strategy by greedily minimizing position errors while neglecting orientation, which drastically reduces the ASR. RTPC and RTPC-9D display the opposite trade-off to TD3, achieving high orientation accuracy at the expense of position precision. Overwhelmed by the massive state-action space, the single-agent PPO fails to decouple system dynamics and struggles to converge. Furthermore, the non-learning strategy lacks adaptability to complex dynamic variations, resulting in significant planning errors.

To effectively mitigate these limitations, the decoupled architecture of DACMP resolves the inherent dynamic dependencies and reduces individual state-action spaces. This design enables the system to achieve stable convergence and effectively avoid the multi-objective optimization trade-off observed in the baselines. Consequently, DACMP demonstrates a more coordinated and balanced learning process.~Attributed to the TESG mechanism, DACMP is effectively constrained by the expert policy during the initial training phase.~This facilitates the acquisition of balanced positive samples for both the manipulator and the base, enabling synergistic multi-objective optimization. Furthermore, the quantitative results in Table \ref{table2} show that DACMP's final performance metrics are significantly superior to those of the expert strategy.~This validates that DACMP not only leverages the non-optimal prior for accelerated convergence but also evolves to outperform the expert strategy.

% includegraphics: [trim=0 9 0 0, clip,] 左 下 右 上
\begin{figure}[!hbt]
  \centering
  \subfloat[Manipulator position and orientation error.]{
    \includegraphics[trim=23 0 62 4, clip, width=0.315\textwidth]{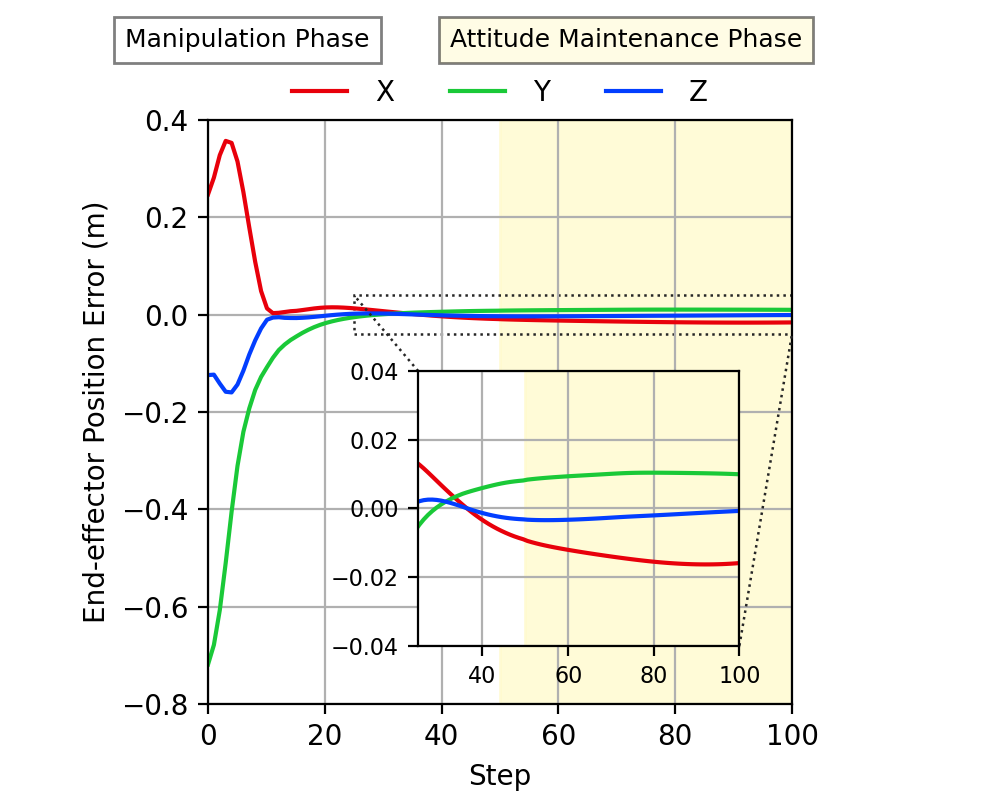}
    \includegraphics[trim=23 0 62 4, clip, width=0.315\textwidth]{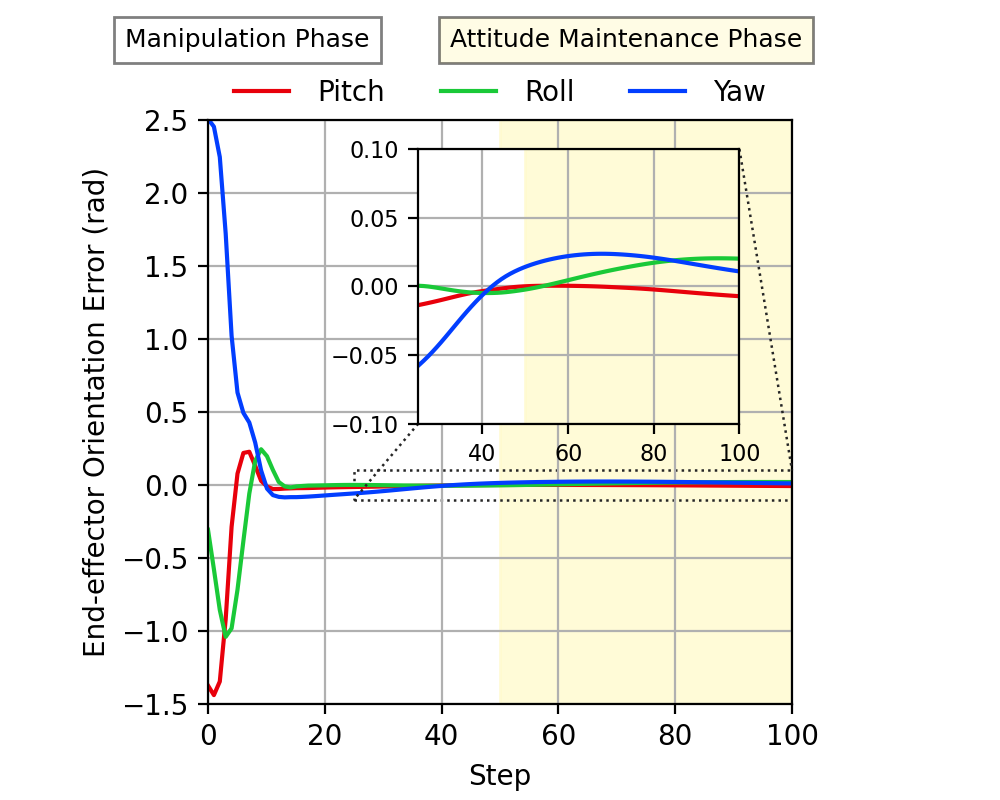}
    \label{fig6_ul}
  }
  % \hfill
  \subfloat[Base attitude error.]{
    \includegraphics[trim=23 0 62 4, clip, width=0.315\textwidth]{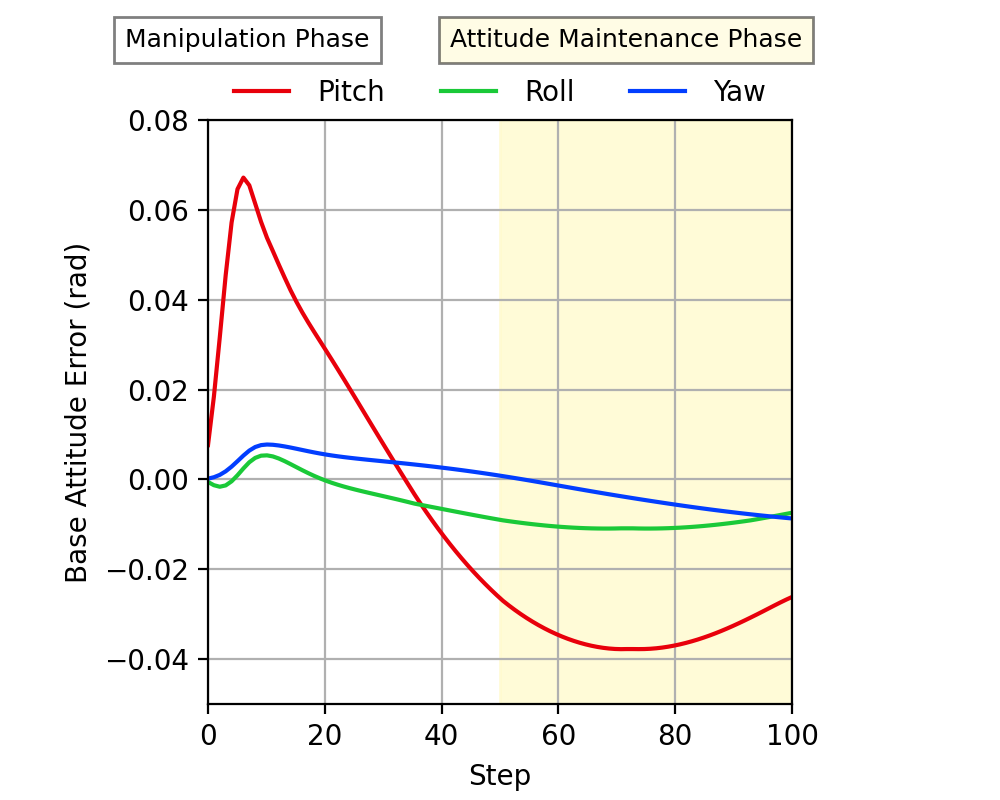}
    \label{fig6_ur}
  } \\
  \subfloat[Manipulator joint angle and angular velocity.]{
    \includegraphics[trim=2 0 2 10, clip, width=0.315\textwidth]{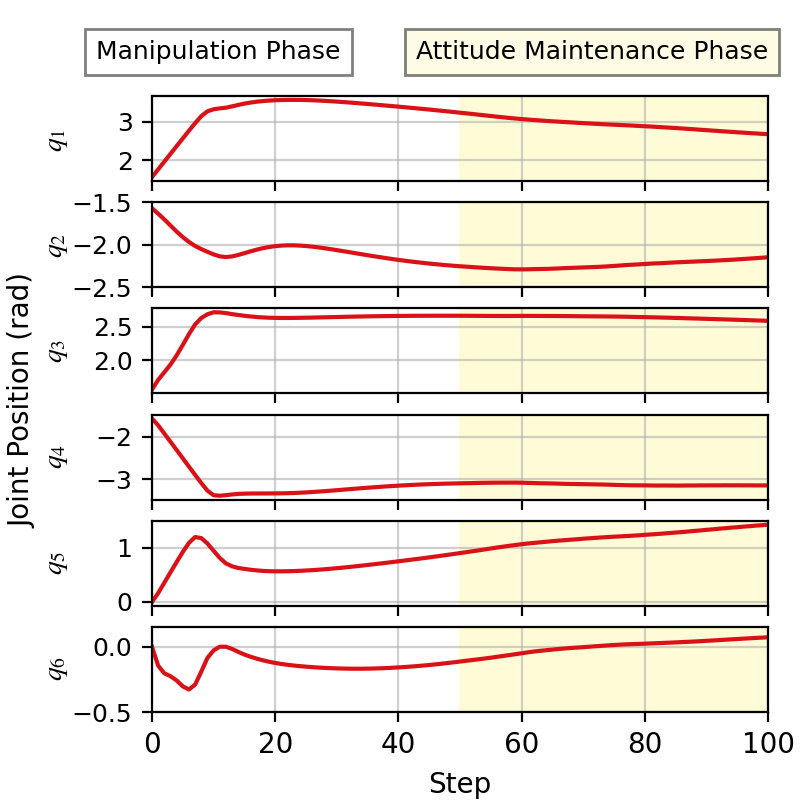}
    \includegraphics[trim=2 0 2 10, clip, width=0.315\textwidth]{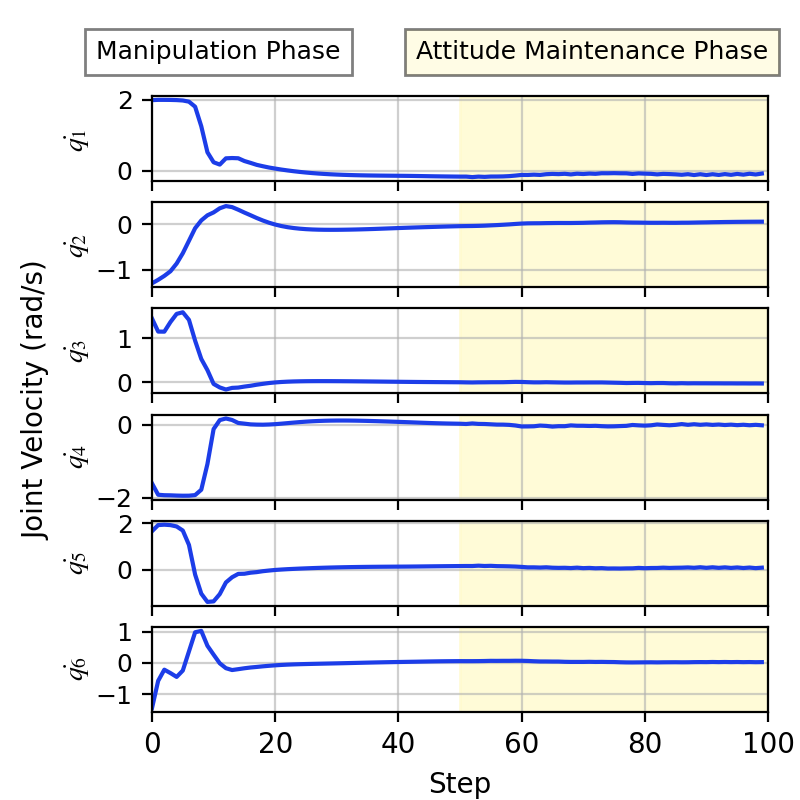}
    \label{fig6_ml}
  }
  % \hfill
  \subfloat[Base control torque.]{
    \includegraphics[trim=2 0 2 10, clip, width=0.315\textwidth]{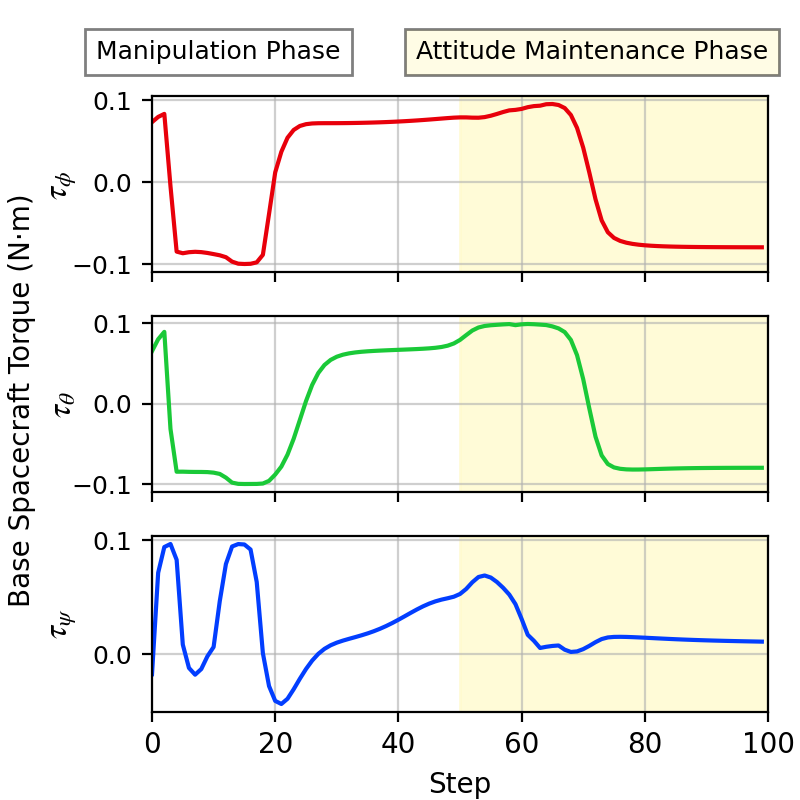}
    \label{fig6_mr}
  } \\
  \subfloat[Spacecraft-manipulator system manipulation planning sequence.
  The red translucent mask indicates the initial base attitude. Step 0: initial configuration, Step 10: rapid approach, Step 30: target pose reached, Step 50: end of motion planning, and Step 100: maintenance completed.]{
    \includegraphics[trim=0 0 0 0, clip, ]{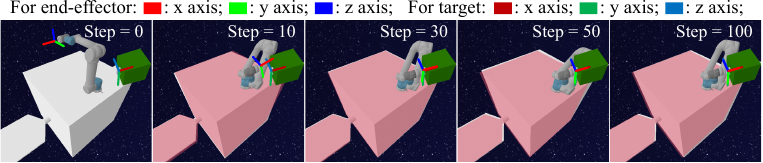}
    \label{fig6_lower}
  }
  \caption{Visual and quantitative performance of manipulation planning using DACMP framework. The white and light-yellow shaded areas represent the manipulation phase and the attitude maintenance phase, respectively.}
  \label{fig6}
\end{figure}

To quantitatively evaluate the performance of the proposed DACMP framework, a challenging task scenario was selected, as shown in Fig.~\ref{fig6}. The target position, defined relative to the base frame $\mathcal{B}$, was set to $(0.002, -0.632, 0.997)$, with the orientation expressed as the quaternion $(0.632, -0.083, 0.330, -0.696)$. In this scenario, the target is located behind the manipulator near the boundary of the dexterous workspace.~This configuration necessitates large-scale manipulator maneuvers, thereby inducing significant dynamic disturbances to the base spacecraft.

Fig.~\ref{fig6_ul} illustrates the convergence precision of the end-effector. The space robot successfully completed the manipulation planning task by the 20th step. From the 30th step onwards, the position and orientation errors stably converged to within 0.01 m and 0.05 rad, respectively, and maintained stable tracking throughout the subsequent attitude maintenance phase. Fig.~\ref{fig6_ur} depicts the attitude response of the base spacecraft over an extended 100-step observation horizon, comprising a manipulation phase that transitions into an attitude maintenance phase at step 50. Due to the strict torque saturation limit of 0.1 $\text{N} \cdot \text{m}$, the Pitch axis exhibited a transient but sharp error increase during the initial 10 steps of rapid maneuvering. However, driven by the DACMP, the system effectively suppressed this early divergence trend. During the maintenance phase, the base agent actively counteracted the residual angular momentum from the preceding maneuvers, successfully arresting the slight increase in absolute Pitch error. From approximately step 70 onwards, the attitude error demonstrated a clear recovery characteristic, exhibiting a steady convergence trend. This distinct suppression and recovery behavior clearly validates the effectiveness of the proposed framework for robust attitude regulation. Fig.~\ref{fig6_ml}~and~\ref{fig6_mr}~further verify the physical feasibility of the generated trajectory. The planned joint angular velocity profiles maintained smooth and continuous variations, with magnitudes asymptotically converging to zero. Furthermore, the corresponding colors intuitively highlight the strict causal relationship between the base control torques in Fig.~\ref{fig6_mr} and the attitude errors in Fig.~\ref{fig6_ur}. Although the base control torques exhibited saturation phases, they show no signs of high-frequency chattering throughout the process. These results demonstrate that DACMP generates physically feasible trajectories and executable control actions compliant with real-world constraints. The corresponding keyframes of the entire sequence, encompassing both the manipulation and attitude maintenance phases, are visualized in Fig.~\ref{fig6_lower}.

\subsection{Ablation studies}
\label{subsec43}
In this section, we conduct ablation studies to identify the contributions of the different components within the DACMP framework.~We first verify the effectiveness of the dual-agent architecture and the expert guidance mechanism, followed by an in-depth analysis of the expert policy and the TESG method.

\subsubsection{Ablation of key components}
\label{subsubsec431}
\begin{figure}[!htb]
  \centering
  \subfloat[Average Success Rate (ASR)]{
    \includegraphics[trim=0 0 34 12, clip, width=0.47\textwidth]{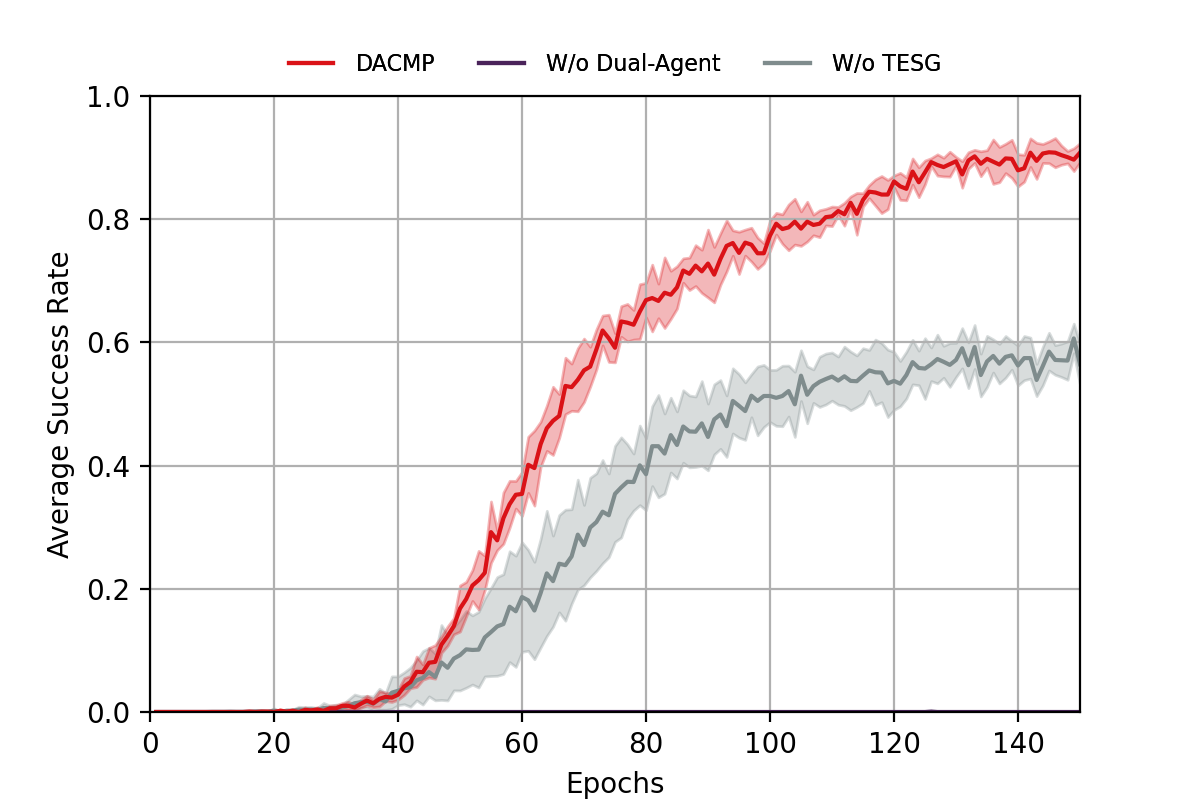}
    \label{fig7_ul}
  }
  % \hfill
  \subfloat[Average Base Attitude Error (ABAE)]{
    \includegraphics[trim=0 0 34 12, clip, width=0.47\textwidth]{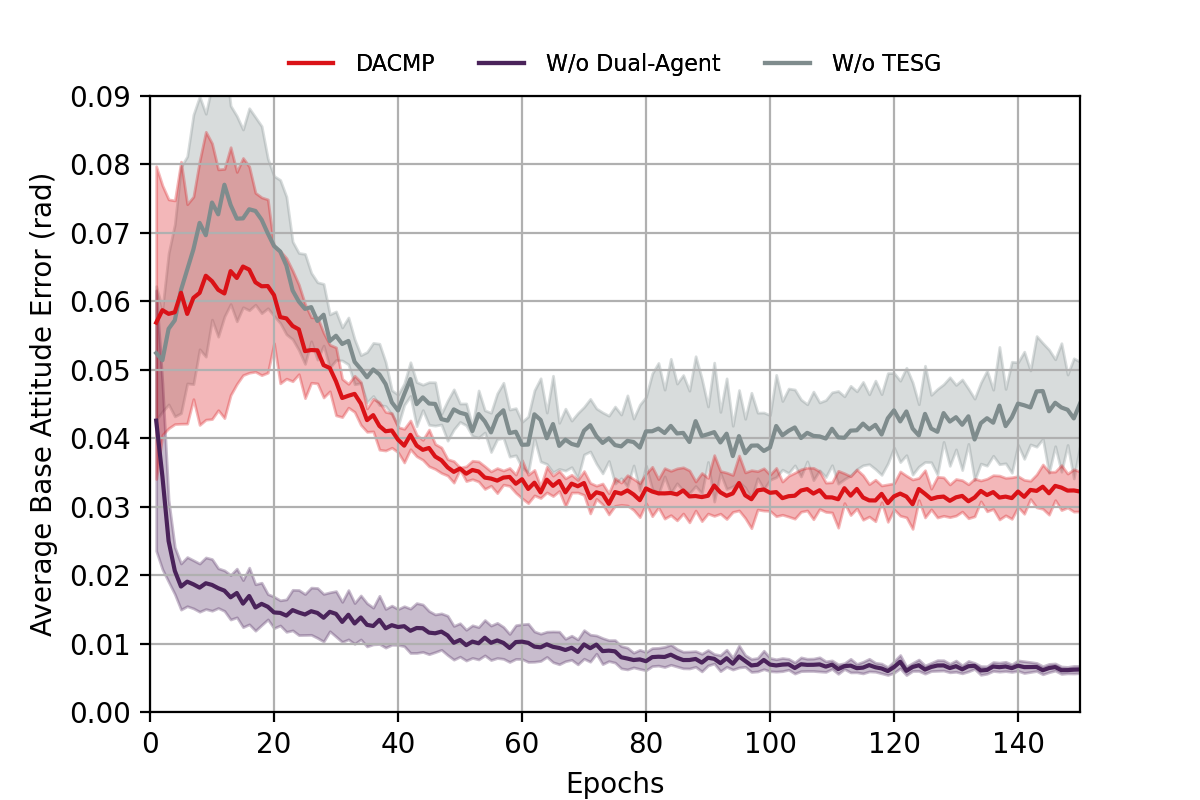}
    \label{fig7_ur}
  } \\
  \subfloat[Average Position Error (APE)]{
    \includegraphics[trim=0 0 34 12, clip, width=0.47\textwidth]{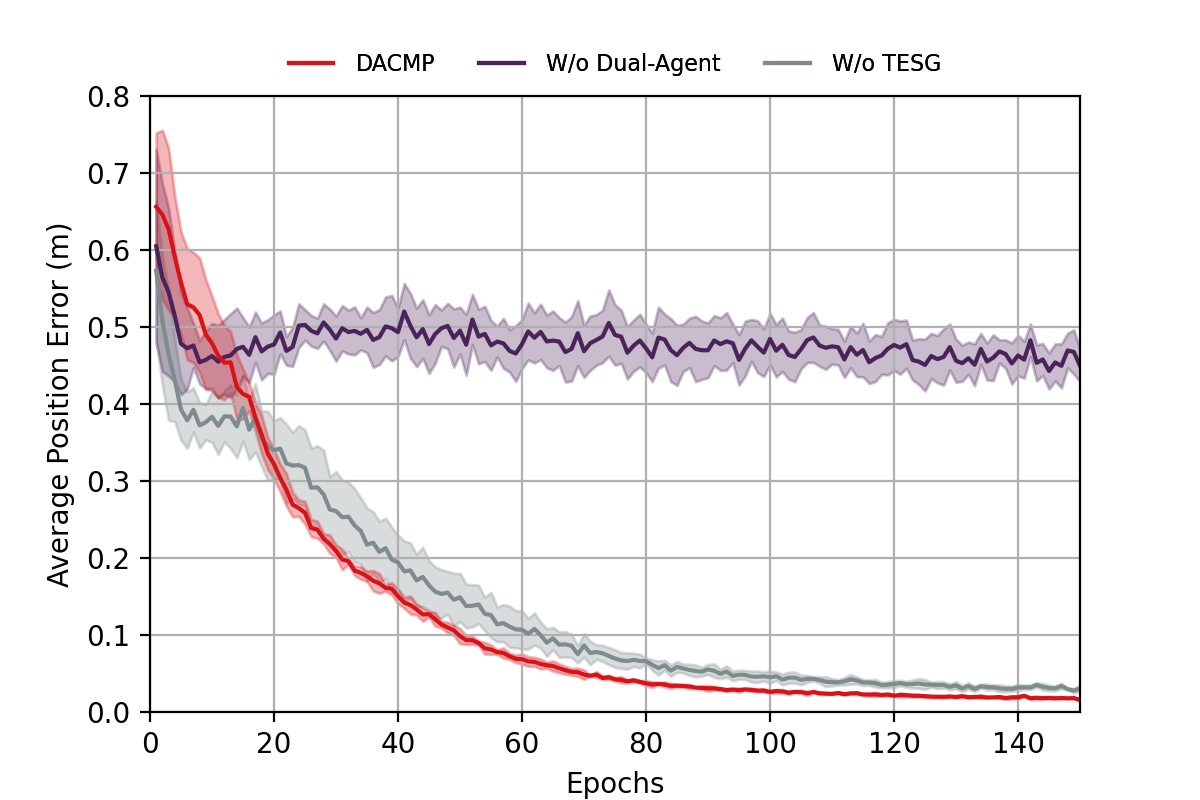}
    \label{fig7_ll}
  }
  % \hfill
  \subfloat[Average Orientation Error (AOE)]{
    \includegraphics[trim=0 0 34 12, clip, width=0.47\textwidth]{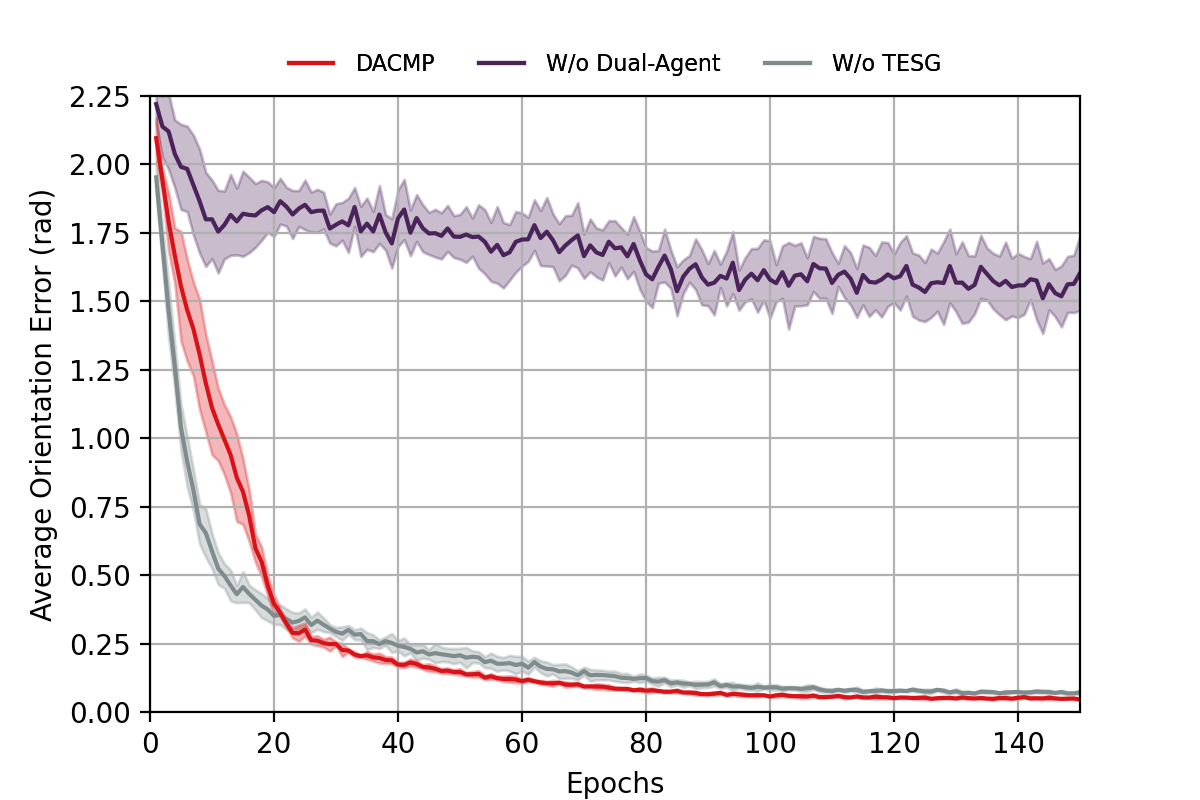}
    \label{fig7_lr}
  }
  \caption{Performance comparison of ablation studies on key components of the DACMP framework.}
  \label{fig7}
\end{figure}

We evaluated the impact of these components by comparing the full DACMP framework against variants lacking the dual-agent architecture and the prior policy guidance, respectively. The comparative curves and quantitative metrics are presented in Fig. \ref{fig7} and Table \ref{table3}.

% includegraphics: [trim=0 9 0 0, clip,] 左 下 右 上

\begin{table}[t]
  \caption{Quantitative results of ablation studies on key components of the DACMP framework.}
  \label{table3}
  \footnotesize
  \centering
  \resizebox{\textwidth}{!}{%
    \begin{tabular}{l c c c c}
      \toprule
      Algorithm & ASR & ABAE [rad] & APE [m] & AOE [rad] \\     % Table header
      \midrule
      DACMP (ours) & $\textbf{0.9080} \pm 0.0154$ & $0.0322 \pm 0.0029$ & $\textbf{0.0156} \pm 0.0005$ & $\textbf{0.0457} \pm 0.0020$ \\
      W/o Dual-agent & $0.0000 \pm 0.0000$ & $\textbf{0.0062} \pm 0.0005$ & $0.4498 \pm 0.0210$ & $1.6027 \pm 0.1366$ \\
      W/o TESG & $0.5645 \pm 0.0330$ & $0.0451 \pm 0.0061$ & $0.0306 \pm 0.0042$ & $0.0718 \pm 0.0074$ \\
      \bottomrule
    \end{tabular}
  }
\end{table}

Fig. \ref{fig7_ul} demonstrates that the dual-agent architecture is vital for task success, as its removal results in a complete breakdown of the manipulation planning task. This deficiency is attributed to optimization conflicts inherent in the single-agent strategy. The agent is driven to maximize the more accessible rewards, specifically those for base attitude, thereby compromising its control over the manipulator. By keeping the manipulator nearly stationary to avoid dynamic disturbances, the single-agent achieves the lowest ABAE reported in Table \ref{table3}, yet entirely fails the task with an ASR of 0. As shown in Fig.~\ref{fig7}, although the single-agent variant rapidly reduces the base attitude error, the end-effector pose fails to converge, ultimately resulting in overall task failure.

The role of prior policy guidance is more complex.~The absence of guidance causes the final ASR to drop below 60\%.~Notably, the unguided variant exhibits a faster decline in APE and AOE than DACMP during the initial training phase. This phenomenon reflects the aggressive nature of unconstrained DRL, where the agent tends to execute large-magnitude, non-smooth maneuvers to maximize immediate rewards.~While yielding short-term reward, this strategy exacerbates base attitude errors, as observed in Fig.~\ref{fig7_ur}, trapping the policy in local optima and limiting long-term convergence. In contrast, DACMP leverages guidance to constrain the exploration process.~Although this results in slower initial adaptation, it ensures trajectory smoothness and action stability, ultimately achieving more robust global convergence.

\subsubsection{Ablation of guidance mechanism}
\label{subsubsec432}
\begin{figure}[!htb]
  \centering
  \subfloat[Average Success Rate (ASR)]{
    \includegraphics[trim=0 0 34 12, clip, width=0.47\textwidth]{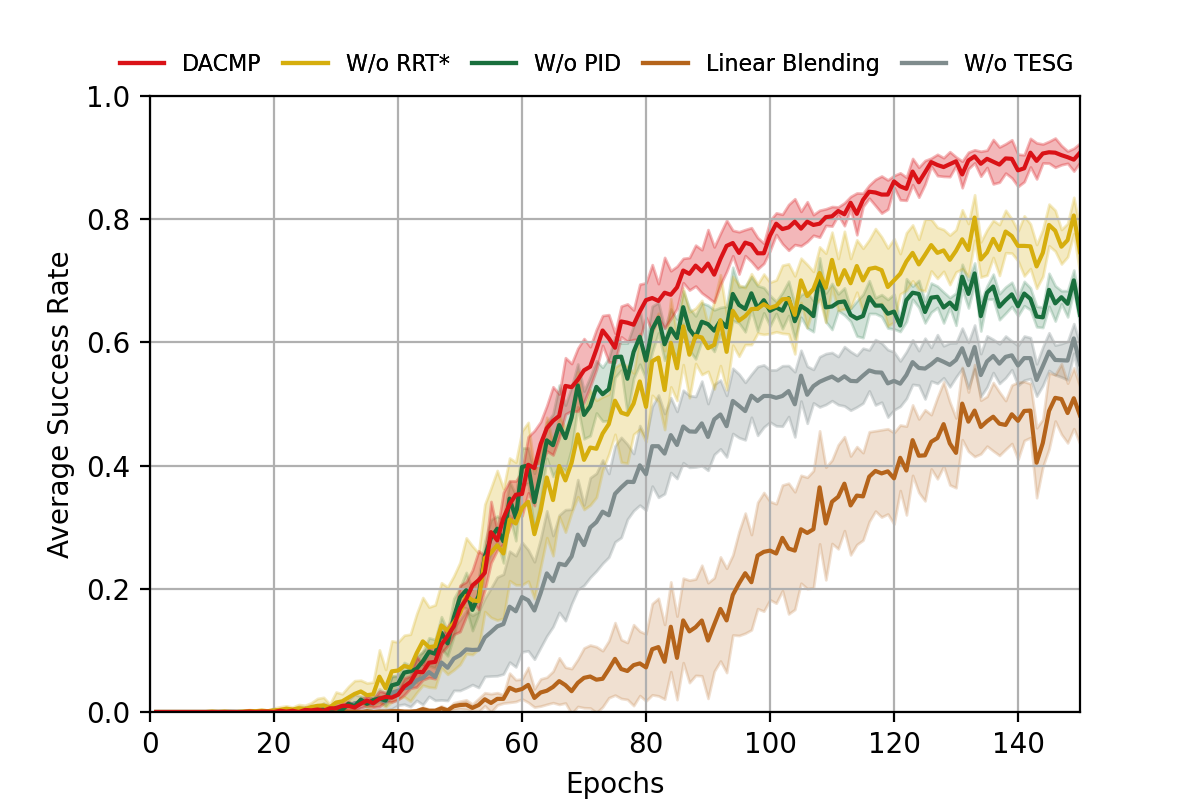}
    \label{fig8_ul}
  }
  % \hfill
  \subfloat[Average Base Attitude Error (ABAE)]{
    \includegraphics[trim=0 0 34 12, clip, width=0.47\textwidth]{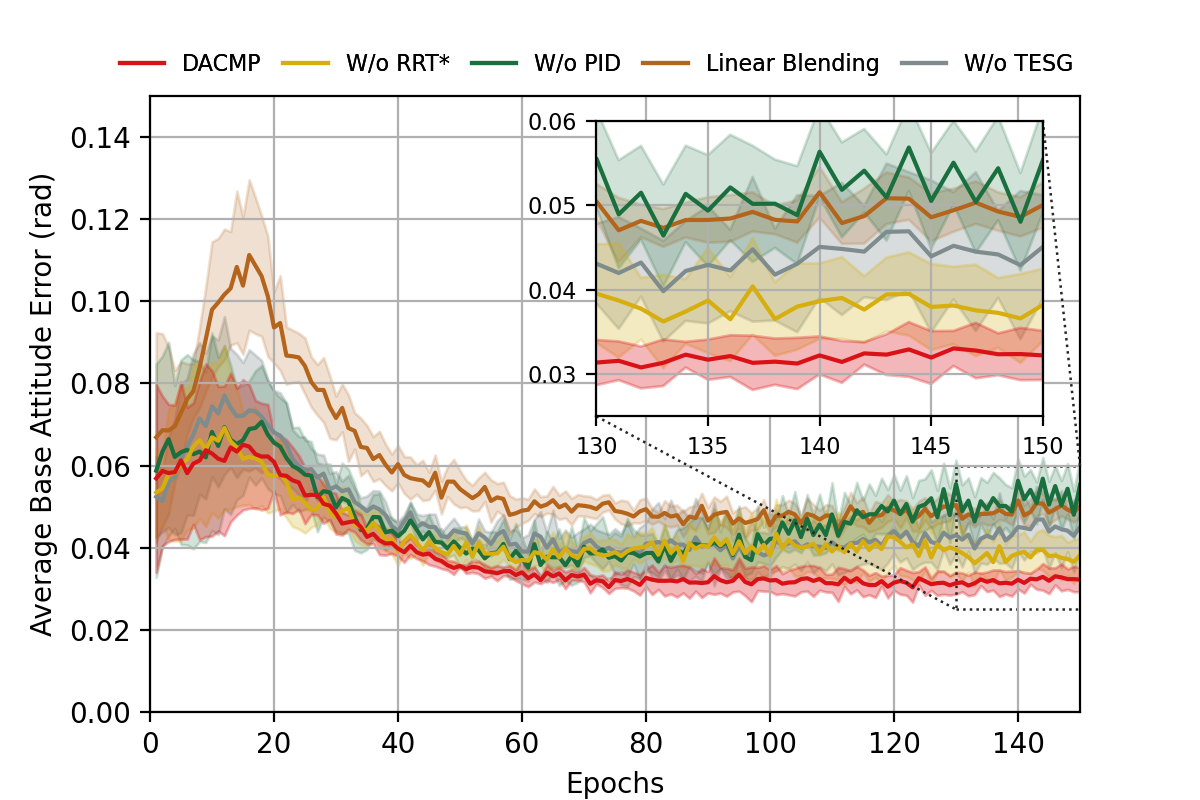}
    \label{fig8_ur}
  } \\
  \subfloat[Average Position Error (APE)]{
    \includegraphics[trim=0 0 34 12, clip, width=0.47\textwidth]{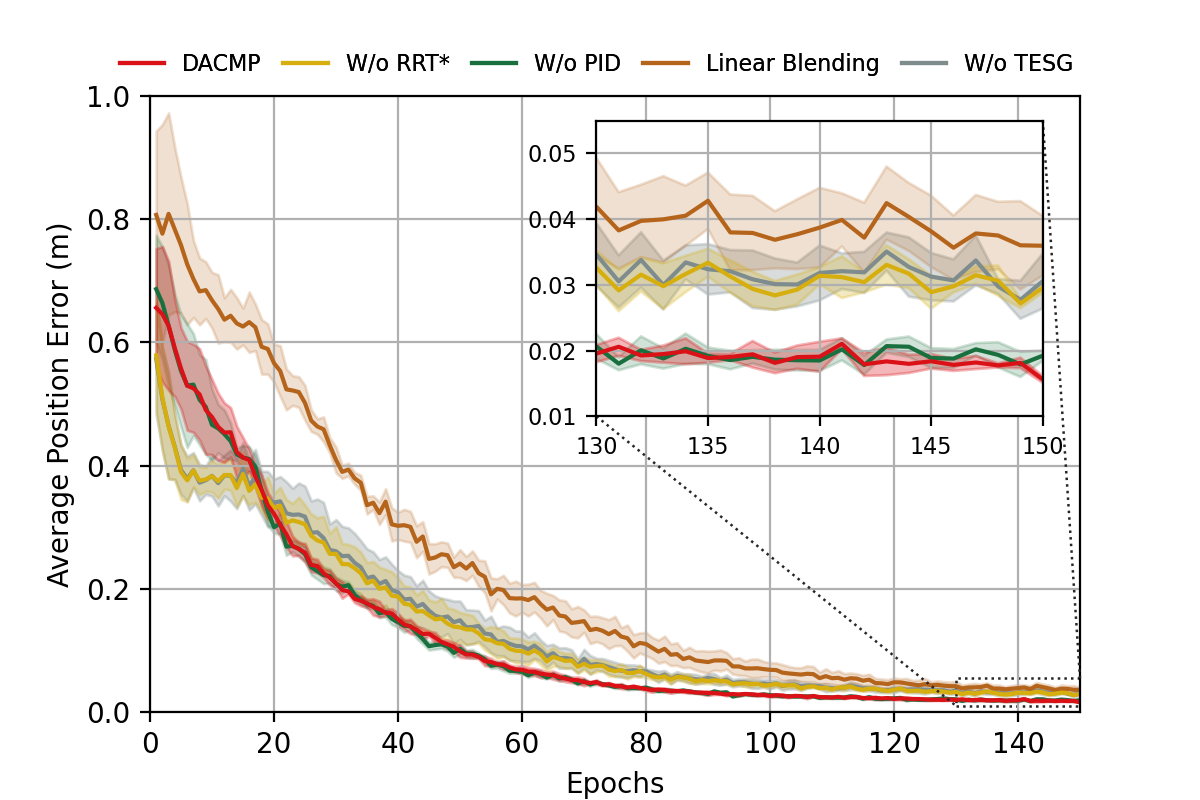}
    \label{fig8_ll}
  }
  % \hfill
  \subfloat[Average Orientation Error (AOE)]{
    \includegraphics[trim=0 0 34 12, clip, width=0.47\textwidth]{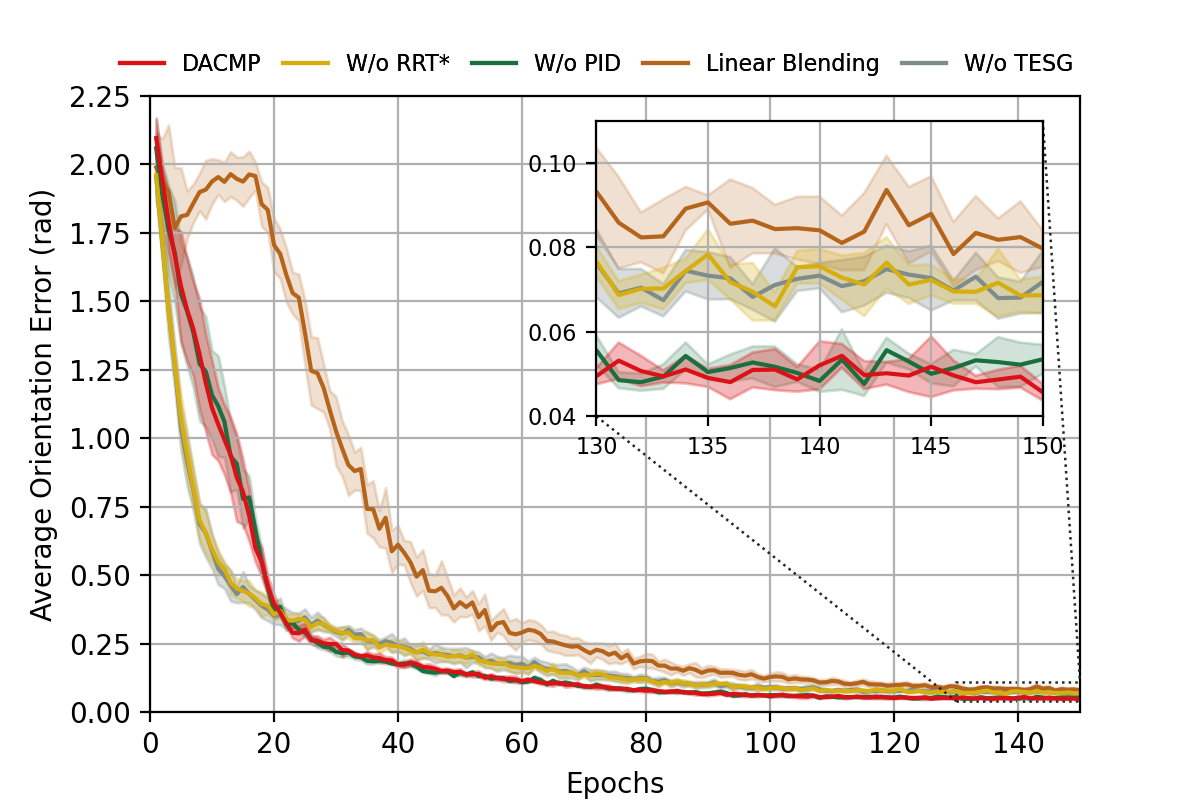}
    \label{fig8_lr}
  }
  \caption{Performance comparison of ablation studies on the guidance mechanism of the DACMP framework.}
  \label{fig8}
\end{figure}

Subsequently, we conducted a detailed analysis to investigate the role of the TESG within DACMP. Specifically, we respectively ablated the RRT* guidance for the manipulator and the PID guidance for the base. Furthermore, to evaluate the blending mechanism, we substituted the proposed TESG method with a traditional linear blending baseline \cite{cao2023reinforcement}. The comparative curves and final metrics are illustrated in Fig. \ref{fig8} and Table \ref{table4}.

% includegraphics: [trim=0 9 0 0, clip,] 左 下 右 上

\begin{table}[b]
  \caption{Quantitative results of ablation studies on the guidance mechanism of the DACMP framework.}
  \label{table4}
  \footnotesize
  \centering
  \resizebox{\textwidth}{!}{%
    \begin{tabular}{l c c c c}
      \toprule
      Algorithm & ASR & ABAE [rad] & APE [m] & AOE [rad] \\     % Table header
      \midrule
      DACMP (ours) & $\textbf{0.9080} \pm 0.0154$ & $\textbf{0.0322} \pm 0.0029$ & $\textbf{0.0156} \pm 0.0005$ & $\textbf{0.0457} \pm 0.0020$ \\
      W/o RRT* & $0.7467 \pm 0.0361$ & $0.0382 \pm 0.0043$ & $0.0296 \pm 0.0008$ & $0.0687 \pm 0.0044$ \\
      W/o PID & $0.6447 \pm 0.0164$ & $0.0553 \pm 0.0063$ & $0.0192 \pm 0.0008$ & $0.0535 \pm 0.0034$ \\
      Linear Blending & $0.4813 \pm 0.0464$ & $0.0500 \pm 0.0027$ & $0.0360 \pm 0.0045$ & $0.0797 \pm 0.0043$ \\
      W/o TESG & $0.5645 \pm 0.0330$ & $0.0451 \pm 0.0061$ & $0.0306 \pm 0.0042$ & $0.0718 \pm 0.0074$ \\
      \bottomrule
    \end{tabular}
  }
\end{table}

It can be clearly seen in Fig. \ref{fig8_ul} that ablating any component of the guidance mechanism leads to performance degradation.~Specifically, the variant without RRT* replicates the "short-sighted" behavior observed in Section \ref{subsubsec431}.~It engages in aggressive initial exploration that leads to local optima traps, thereby limiting long-term convergence, as illustrated in Fig.~\ref{fig8_ll} and~\ref{fig8_lr}.~The impact of removing PID guidance is even more pronounced on base stability, causing initial error degradation followed by divergence and severe high-frequency jitter, as shown in Fig.~\ref{fig8_ur}. These results confirm that a complete guidance mechanism is indispensable for robust task convergence.

The variant using Linear Blending exhibits distinctive learning dynamics, particularly in Fig. \ref{fig8_ur} and \ref{fig8_lr}. Notably, during the guidance phase, both the base attitude and end-effector orientation errors show an anomalous upward trend. A clear convergence trend appears only after the linear blending guidance completely fades out. This phenomenon highlights the advantage of TESG in mitigating the optimization conflict inherent in rigid linear blending strategies. Unlike blending methods that impose a mixed action, TESG dynamically regulates the probability of expert intervention. This mechanism prevents the agent from learning to merely compensate for the superimposed prior signal, ensuring it solves the task independently.

A more notable observation is the distinct alignment of specific ablation curves regarding end-effector pose errors. Referring to Fig. \ref{fig8_ll} and \ref{fig8_lr}, the trajectory of the variant without RRT* aligns almost perfectly with that of the completely unguided baseline. Conversely, the variant without PID remains virtually identical to the full DACMP method in these same metrics, despite the lack of base guidance.~We attribute this phenomenon to the effective decoupling capability of the dual-agent architecture.~It demonstrates that the presence or absence of guidance for one specific control objective does not significantly affect the optimization process of the other. This independence strongly validates the architecture's efficacy in handling multi-objective coupled tasks.

\subsection{Robustness analysis}
\label{subsec44}
In this section, we assess the robustness of the DACMP framework against typical space disturbances and uncertainties, including environmental factors, system constraints, and model uncertainties. Notably, the model was trained exclusively in a nominal environment and is evaluated directly under these disturbances without any fine-tuning. Each disturbance is introduced into the simulation independently to decouple its effect from others. Moreover, the aggregated results serve as rigorous statistical evidence for the empirical stabilization of the proposed framework.

\subsubsection{Environmental disturbances}
\label{subsubsec441}
\begin{figure}[!htb]
  \centering
  \subfloat[Impact of target spin rates.]{
    \includegraphics[trim=10 9 12 10, clip, width=0.47\textwidth]{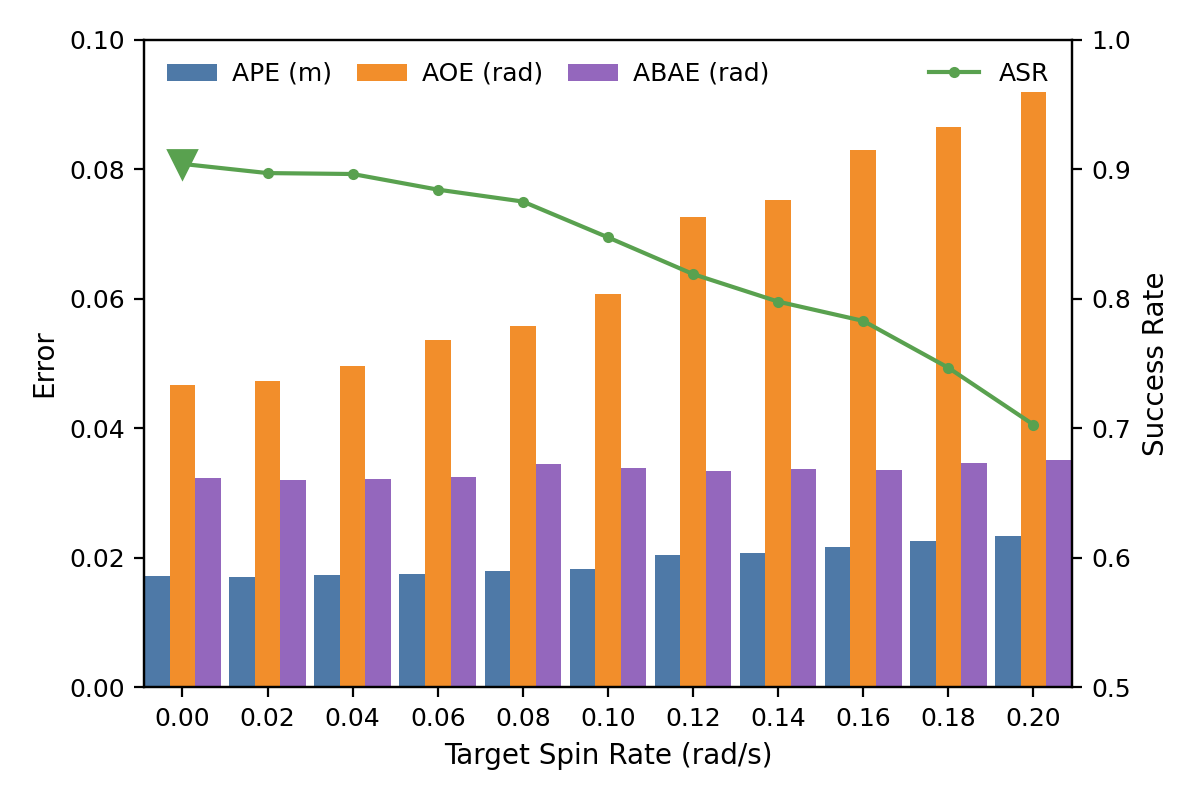}
    \label{fig9_l}
  }
  % \hfill
  \subfloat[Impact of external base disturbance torques.]{
    \includegraphics[trim=10 9 12 10, clip, width=0.47\textwidth]{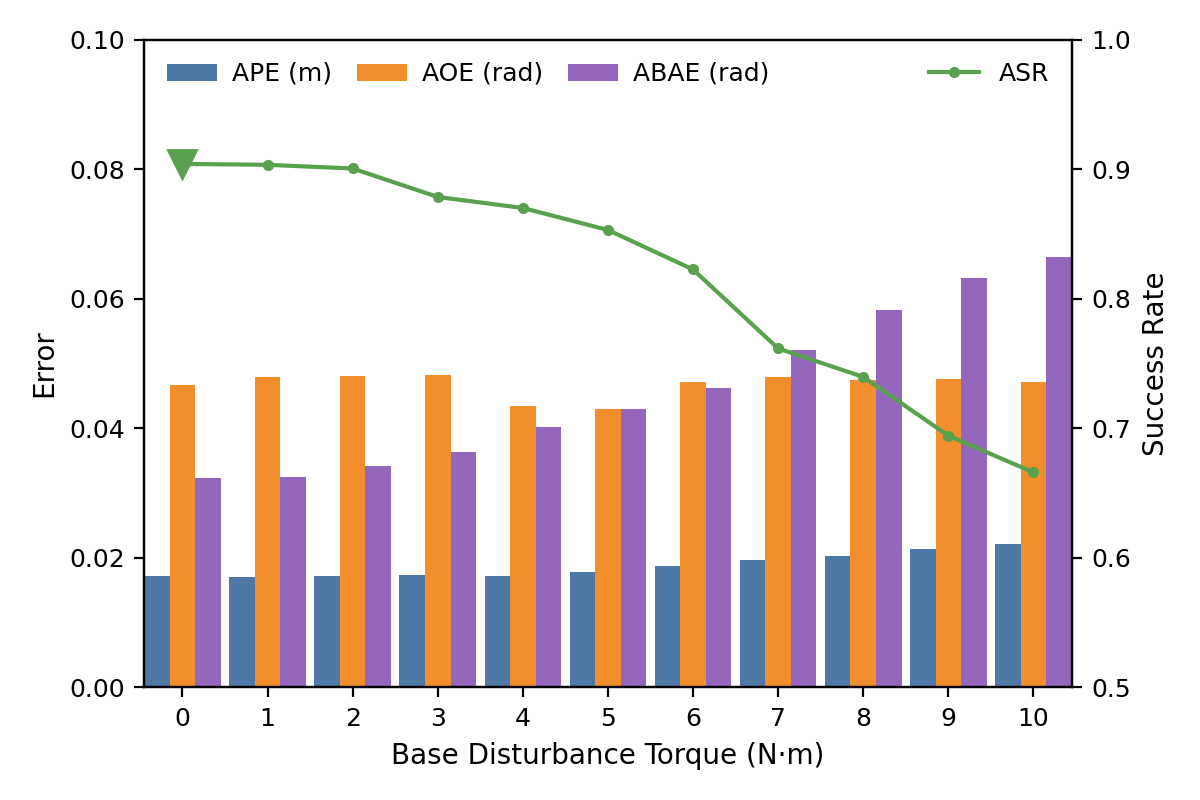}
    \label{fig9_r}
  }
  \caption{Robustness evaluation of the DACMP framework under environmental disturbances. In each subplot, the bars represent the final pose errors (left y-axis), the green curve indicates the task success rate (right y-axis), and the triangle denotes the original ASR. Each data point represents the average value over 5 random seeds, with 5000 simulation episodes per seed.}
  \label{fig9}
\end{figure}

Space environmental disturbances inevitably disrupt the static equilibrium of both the target and the space robot system. Consequently, we simulate two representative perturbations, including target spin and external disturbance torque applied to the base. Fig. \ref{fig9} depicts the robustness performance under these conditions. In the figure, the x-axis represents the disturbance magnitude, while the direction of the spin or torque is randomized for each simulation episode.

% includegraphics: [trim=0 9 0 0, clip,] 左 下 右 上

Fig.~\ref{fig9_l}~shows that DACMP exhibits strong robustness against spin rates below $0.10$ rad/s.~Beyond this threshold, performance degrades gracefully.~Notably, orientation error proves more sensitive than position error, reflecting the inherent difficulty of high-speed pose tracking.~Crucially, the base attitude error remains consistently low regardless of target spin, confirming the effective decoupling of the dual-agent architecture. Even under the extreme condition of $0.20$ rad/s, DACMP maintains a viable success rate of approximately $70\%$.

Compared to target spin, the system proves more sensitive to direct base disturbance, as shown in Fig. \ref{fig9_r}. Performance begins to decline noticeably when the external torque exceeds $3$ $\text{N} \cdot \text{m}$. Due to limited actuation capacity, the base struggles to counteract such high-magnitude impulses, resulting in accumulating attitude errors.~Through dynamic coupling, the base instability propagates directly to the end-effector, causing significant orientation fluctuations.~Despite these hardware constraints, DACMP still achieves a success rate of approximately 0.66 even under $10$ $\text{N} \cdot \text{m}$ disturbances. This resilience suggests the manipulator agent compensates for base instability, preventing catastrophic failure.

\subsubsection{System constraints and faults}
\label{subsubsec442}
% includegraphics: [trim=0 9 0 0, clip,] 左 下 右 上
\begin{figure}[!b]
  \centering
  \subfloat[Impact of observation delays.]{
    \includegraphics[trim=10 9 12 10, clip, width=0.47\textwidth]{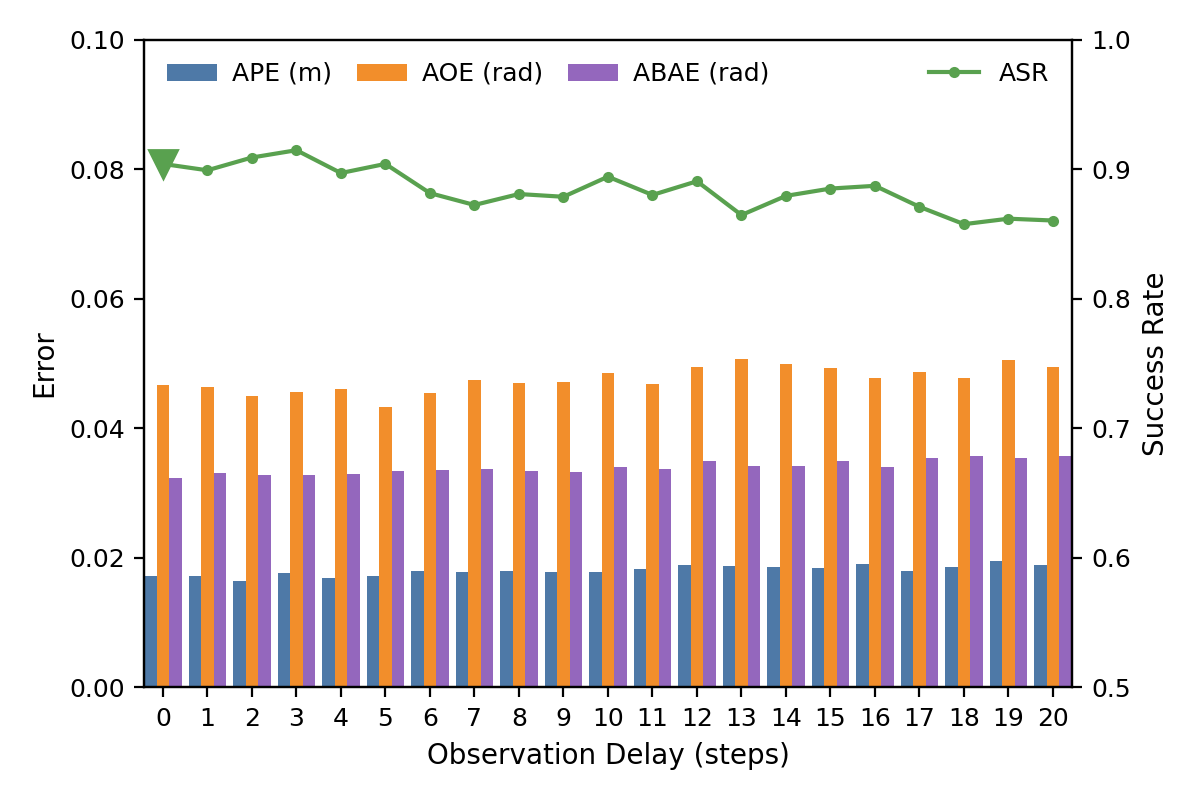}
    \label{fig10_l}
  }
  % \hfill
  \subfloat[Impact of action delays.]{
    \includegraphics[trim=10 9 12 10, clip, width=0.47\textwidth]{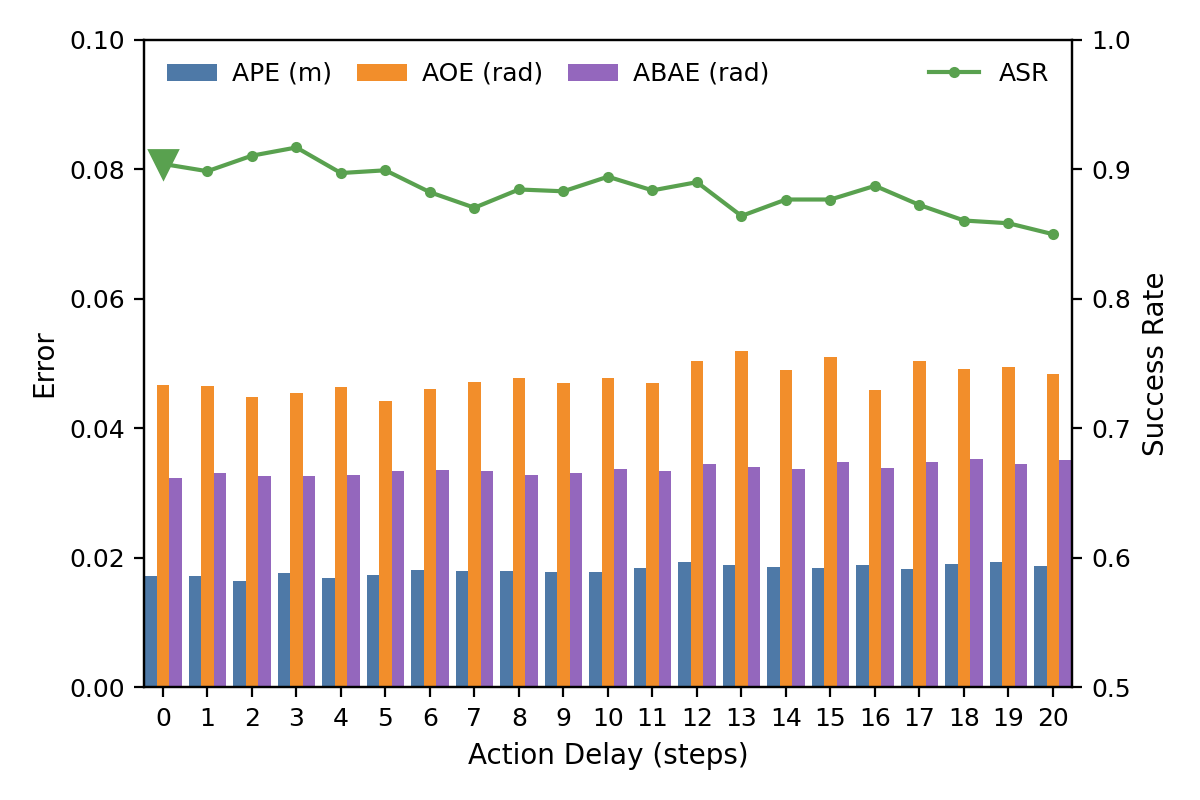}
    \label{fig10_r}
  }
  \caption{Robustness evaluation of the DACMP framework under observation and action delays.}
  \label{fig10}
\end{figure}

Beyond external environmental disturbances, internal system constraints and potential faults pose significant challenges to the manipulation planning task. In this part, we evaluate the framework against three critical issues, namely transmission delays, actuator efficiency loss, and control saturation. To simulate the transmission delays, we implemented a stochastic delay mechanism starting from the second simulation step, where the observations or actions are randomly replaced by their values from the preceding step. The robustness evaluation results of observation and action delays are illustrated in Fig. \ref{fig10}.

The DACMP demonstrates remarkable resilience against signal transmission delays, proving equally insensitive to both observation and action delays. Surprisingly, even under severe latency conditions, the system shows no statistically significant performance degradation. The ASR curves fluctuate within a narrow margin, a behavior we attribute to simulation stochasticity rather than systemic failure. This robustness suggests that, unlike model-based approaches sensitive to synchronization, the learning-based DACMP possesses superior generalization. By directly mapping states to actions, the agents inherently tolerate temporal noise, ensuring effective control even with intermittent feedback.

\begin{figure}[!h]
  \centering
  \subfloat[Impact of base spacecraft efficiency loss.]{
    \includegraphics[trim=10 9 12 10, clip, width=0.47\textwidth]{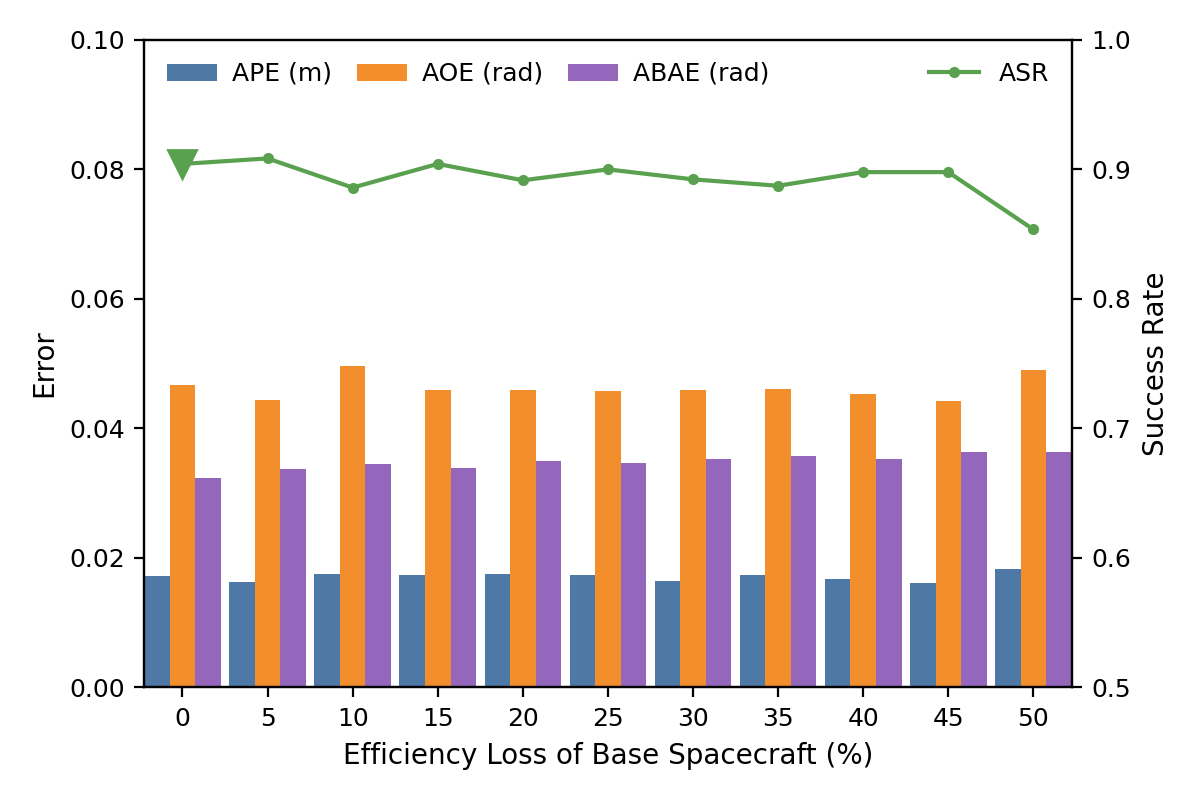}
    \label{fig11_l}
  }
  % \hfill
  \subfloat[Impact of manipulator efficiency loss.]{
    \includegraphics[trim=10 9 12 10, clip, width=0.47\textwidth]{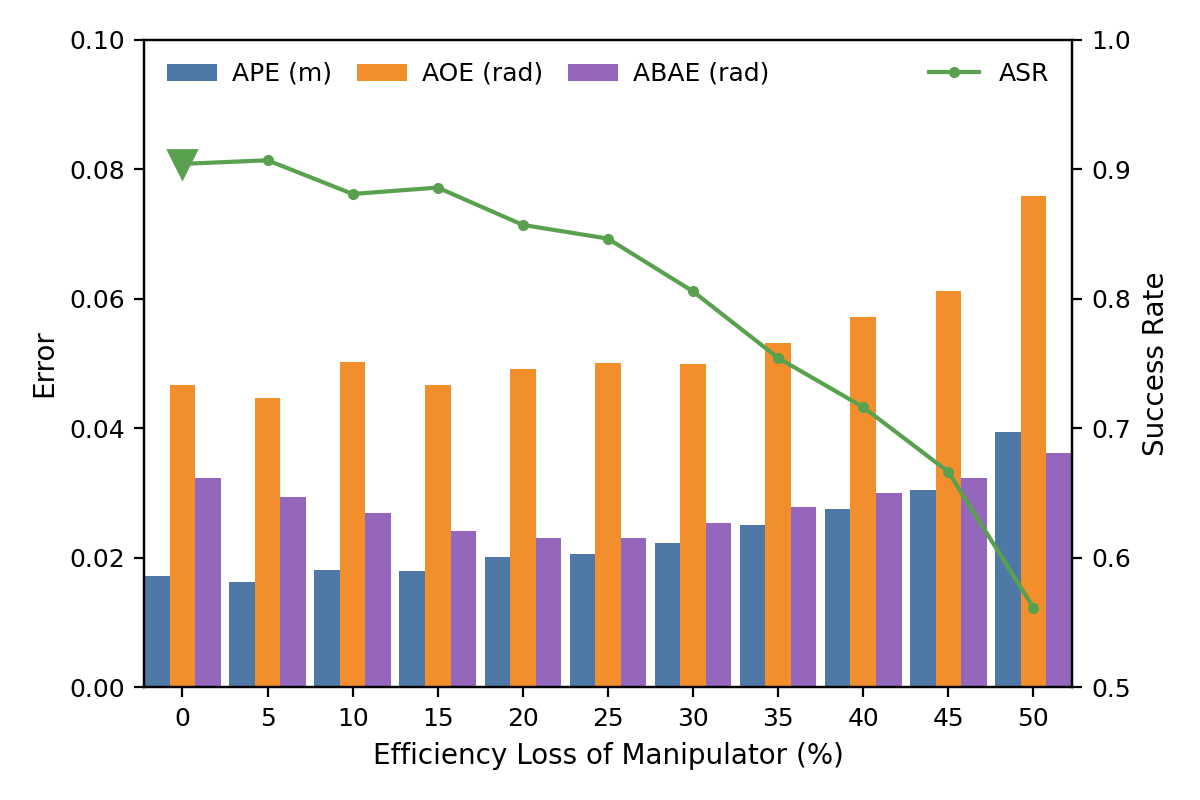}
    \label{fig11_r}
  }
  \caption{Robustness evaluation of the DACMP framework under base spacecraft and manipulator efficiency loss.}
  \label{fig11}
\end{figure}

Furthermore, to rigorously test the system's dynamic compensation and closed-loop adaptability against severe hardware degradation, we evaluate the framework under actuator efficiency loss. This fault represents a proportional attenuation of the desired manipulator joint velocity and base output torque, as presented in Fig.~\ref{fig11}.

The evaluation results confirm that DACMP exhibits remarkable resilience against base actuator degradation. As shown in Fig.~\ref{fig11_l}, the ASR remains stable near 0.9 for base efficiency losses up to 45\%, dropping only slightly to 0.85 at a 50\% loss. In contrast, Fig.~\ref{fig11_r} reveals the system's higher sensitivity to manipulator actuator degradation. The ASR remains steady for loss levels up to 20\%. However, beyond this threshold, the manipulator struggles to reach the target pose, causing a nearly linear ASR drop to roughly 0.55 under a 50\% loss. Notably, ABAE exhibits a non-monotonic trend, dipping before rising beyond a 20\% manipulator efficiency loss. This suggests an implicit coordination between the agents. Moderate arm degradation inadvertently reduces dynamic disturbances on the base. However, a severe loss forces the base into aggressive compensatory maneuvers, ultimately increasing its own error.

Finally, we investigate the impact of actuator saturation. In practice, attitude control actuators, such as reaction wheels, are constrained by limited momentum storage capacity. As these actuators approach saturation, their ability to generate control torques diminishes significantly. To evaluate this condition, we simulate a three-axis attitude control system with a capacity of 3 $\text{N} \cdot \text{m} \cdot \text{s}$, analyzing mission performance under varying initial momentum saturation levels.

% includegraphics: [trim=0 9 0 0, clip,] 左 下 右 上
\begin{figure}[!htb]
  \centering
  \subfloat[Impact of initial momentum saturation.]{
    \includegraphics[trim=10 9 12 10, clip, width=0.47\textwidth]{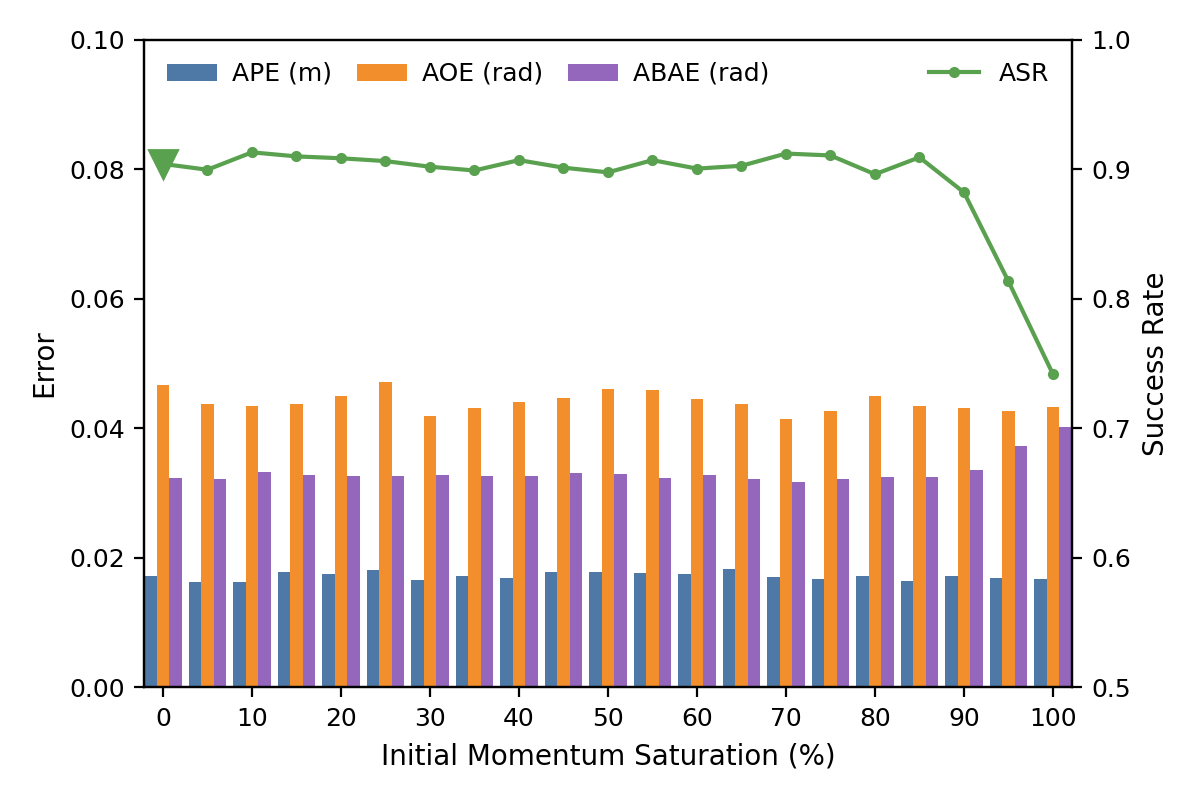}
    \label{fig12_l}
  }
  % \hfill
  \subfloat[Impact of base mass variations.]{
    \includegraphics[trim=10 9 12 10, clip, width=0.47\textwidth]{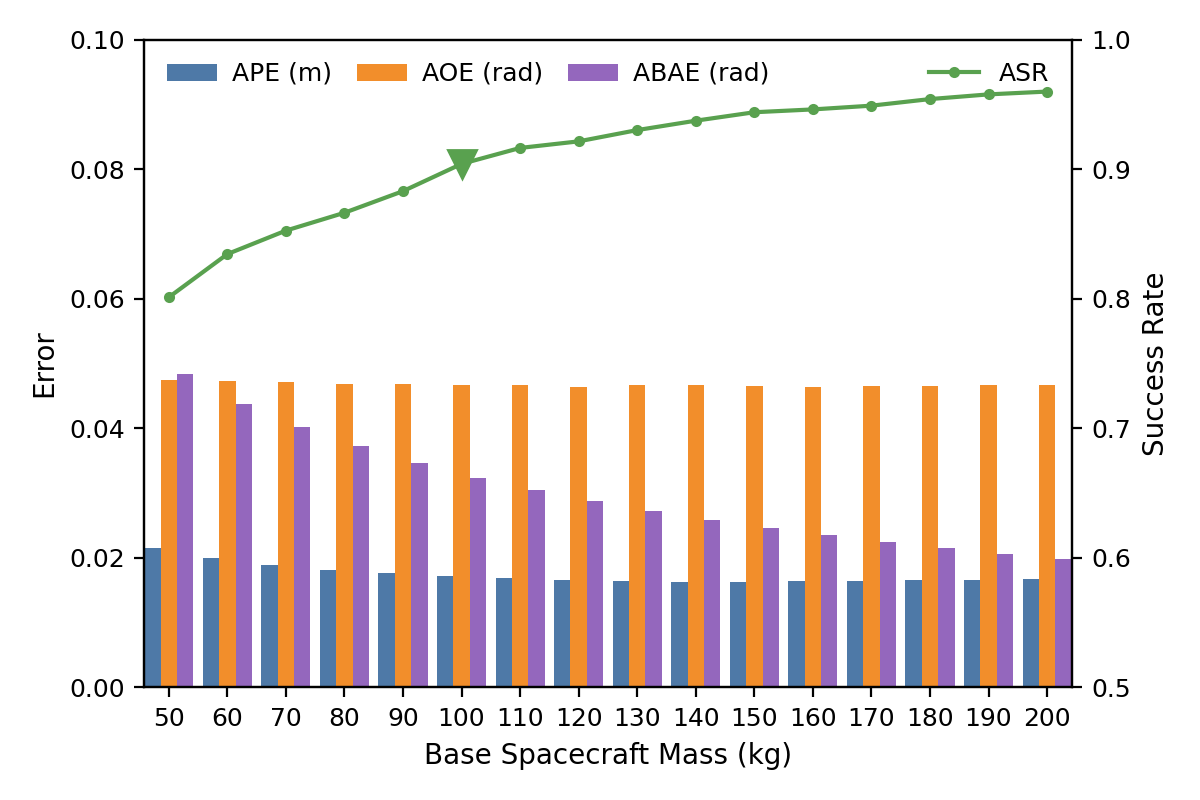}
    \label{fig12_r}
  }
  \caption{Robustness evaluation of the DACMP framework under actuator saturation and base mass variations.}
  \label{fig12}
\end{figure}

Fig.~\ref{fig12_l}~demonstrates that the system maintains consistent performance even with initial momentum saturation up to 85\%.~This stability is physically grounded, since the remaining momentum capacity of 0.45 $\text{N} \cdot \text{m} \cdot \text{s}$ is sufficient to cover the typical angular impulse required for the manipulation planning task. However, as the saturation exceeds 90\%, the success rate exhibits a sharp linear decline. This degradation results from the depletion of momentum margin, which deprives the base of control authority to counteract disturbances. Notably, even under complete saturation, DACMP still achieves a success rate of approximately 0.75. This resilience proves that the manipulator agent can partially compensate for the failure of the base spacecraft, ensuring mission success despite hardware limitations.

\subsubsection{Model and perception uncertainties}
\label{subsubsec443}
Finally, we address the challenges posed by model mismatches and perception uncertainties. In real-world missions, inertial properties often fluctuate due to fuel consumption or payload variations. Concurrently, state estimation is inevitably corrupted by sensor noise.~To assess robustness against these factors, we first evaluate the impact of base mass variations, as illustrated in Fig.~\ref{fig12_r}. The system exhibits distinct behaviors depending on the mass variation direction. Increased base mass enhances inertia, naturally suppressing disturbances and improving accuracy. Conversely, mass reduction poses a critical challenge as a lighter base is inherently less stable against coupling torques. Despite this structural vulnerability, DACMP maintains satisfactory performance even with a 50\% mass reduction, achieving an ASR of approximately 0.79. This proves the algorithm's adaptability to severe model mismatches, preventing divergence even with compromised base stability.

\begin{figure}[!hbt]
  \centering
  \subfloat[Impact of initial position observation error.]{
    \includegraphics[trim=10 9 12 10, clip, width=0.47\textwidth]{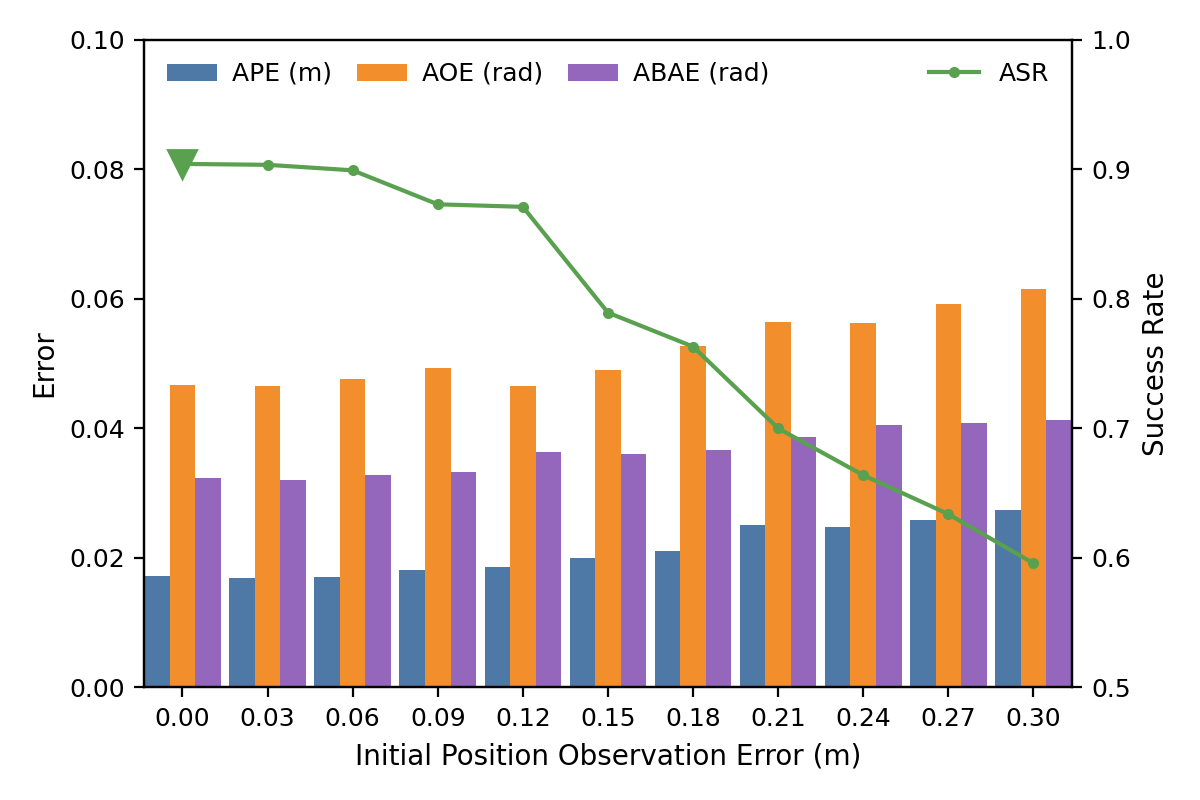}
    \label{fig13_l}
  }
  % \hfill
  \subfloat[Impact of initial orientation observation error.]{
    \includegraphics[trim=10 9 12 10, clip, width=0.47\textwidth]{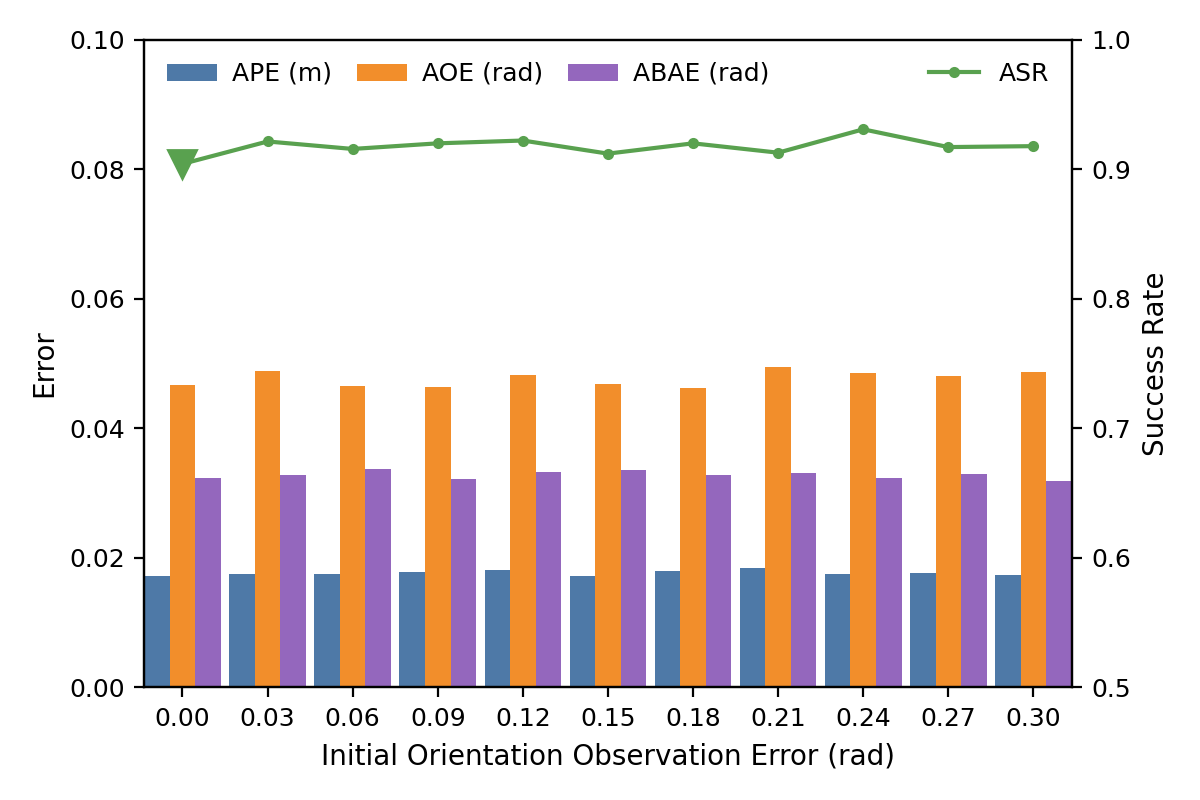}
    \label{fig13_r}
  }
  \caption{Robustness evaluation of the DACMP framework under initial target observation errors.}
  \label{fig13}
\end{figure}

Finally, we investigate the impact of perception uncertainties by introducing initial observation biases. Specifically, position or orientation observation errors are superimposed on the target state at the initialization phase. To mimic the typical convergence behavior of visual estimation algorithms, these errors are designed to decay linearly, vanishing completely after the 30th simulation step. The impact of these varying initial error magnitudes on DACMP is shown in Fig. \ref{fig13}.

% includegraphics: [trim=0 9 0 0, clip,] 左 下 右 上

As indicated in Fig. \ref{fig13_l}, the system maintains robust performance when the initial position observation error is within 0.12 m. However, as the error magnitude increases further, a gradual degradation is observed in both end-effector positioning accuracy and base attitude stability. Specifically, the ASR drops below 0.70 when the initial position error surpasses 0.21 m.~This deterioration occurs because large initial deviations force the manipulator to perform aggressive maneuvers in the early phase. These rapid motions generate excessive coupling torques, which destabilize the base and risk mission failure.~In sharp contrast, as illustrated in Fig.~\ref{fig13_r}, DACMP demonstrates exceptional robustness against orientation observation errors, maintaining high success rates throughout the tested range. We attribute this insensitivity to the kinematic structure, where orientation corrections primarily rely on the low-inertia wrist joints, generating minimal coupling disturbances to the base compared to the heavy proximal joints.

\section{Conclusions}
\label{sec5}
In this paper, we present the DACMP framework to establish a novel decoupled planning paradigm for spacecraft-manipulator systems. By explicitly isolating manipulation and base stabilization, this architecture resolves severe dynamic coupling, clearly outperforming traditional single-agent approaches. Furthermore, we propose the Timestep-level Expert Switching Guidance (TESG) method to tackle the exploration challenge in high-dimensional spaces.~Unlike linear blending, TESG prevents the agent from merely learning residual compensation, ensuring it develops the capability to solve tasks independently.~Extensive simulations confirm that DACMP outperforms baselines in both success rate and precision.~Notably, the agent demonstrates an implicit understanding of system dynamics, actively compensating for hardware constraints and coupling effects. This implicit physical insight ensures robust performance against signal delays, model mismatches, and perception uncertainties.

Future work will focus on bridging the gap between simulation and real-world deployment. We aim to incorporate visual encoders to achieve end-to-end planning directly from raw images \cite{wang2024vision}, replacing the current reliance on ground-truth states. Additionally, we plan to transition from kinematic velocity control to dynamic torque control \cite{sampath2024intelligent}. This shift will better address real-world physical constraints, paving the way for practical on-orbit applications.

\appendix
\section{Hyperparameters of learning-based algorithms}
\label{appA}
In this appendix, we provide the detailed hyperparameters of all learning-based algorithms used in this paper for reproducibility, as listed in Table \ref{tableap1}.

\begin{table}[!h]
  \caption{Hyperparameters of learning-based algorithms.}
  \label{tableap1}
  \footnotesize
  \centering
  \resizebox{\textwidth}{!}{%
    \begin{tabular}{l c c c c c}
    %% Tabular cells are separated by &
      \toprule
      Hyperparameters & DACMP & TD3 & RTPC & RTPC-9D & PPO \\     % Table header
      \midrule
      Actor \& Critic Networks & -- & -- & -- & (256, 256, 128) & (128, 64, 64)\\
      \quad Manipulator & (256, 256, 128) & (256, 256, 128) & (256, 256, 128) & -- & --\\
      \quad Base Spacecraft & (32, 128, 32) & -- & -- & -- & -- \\
      Activation function & Tanh & Tanh & Tanh & Tanh & Tanh \\
      Learning rate of actor & $2 \times 10 ^{-4}$ & $2 \times 10 ^{-4}$ & $2 \times 10 ^{-4}$ & $2 \times 10 ^{-4}$ & $3 \times 10 ^{-4}$ \\
      Learning rate of critic & $1 \times 10 ^{-4}$ & $1 \times 10 ^{-4}$ & $1 \times 10 ^{-4}$ & $1 \times 10 ^{-4}$ & $3 \times 10 ^{-4}$ \\
      Optimizer & Adam & Adam & Adam & Adam & Adam \\
      Experience buffer size $(C)$ & $8 \times 10 ^4$ & -- & $8 \times 10 ^4$ & $8 \times 10 ^4$ & $8 \times 10 ^4$ \\
      Sample minibatch size $(N)$ & $8 \times 10 ^3$ & -- & $8 \times 10 ^3$ & $8 \times 10 ^3$ & $8 \times 10 ^3$ \\
      Replay buffer size & -- & $1 \times 10 ^6$ & -- & -- & -- \\
      Batch size & -- & 64 & -- & -- & -- \\
      Training episodes $(M)$ & $2.4 \times 10 ^5$ & $4 \times 10 ^5$ & $4 \times 10 ^5$ & $4 \times 10 ^5$ & $4 \times 10 ^5$ \\
      Steps per episode $(T)$ & 50 & 50 & 50 & 50 & 50 \\
      Discount factor $(\gamma)$ & 0.96 & 0.96 & 0.94 & 0.94 & 0.96 \\
      TD decay factor $(\lambda)$ & 0.95 & -- & 0.95 & 0.95 & 0.95 \\
      PPO clipping factor $(\epsilon)$ & 0.1 & -- & 0.2 & 0.2 & 0.2 \\
      Gradient update steps $(K_\text{update})$ & 90 & -- & 45 & 45 & 45 \\
      HER epochs & 70 & -- & 100 & 100 & -- \\
      Guidance epochs $(k_\text{g})$ & 15 & -- & -- & -- & -- \\
      \bottomrule
    \end{tabular}
  }
\end{table}

\section{Complete training pseudocode of DACMP framework}
\label{appB}
In this appendix, we present the complete training procedure for the proposed DACMP framework. Algorithm \ref{alg} details the interaction between the off-policy learning strategy and the TESG mechanism. Concretely, this off-policy workflow strictly separates data collection from optimization. The agents first fill their experience buffers, then execute multiple network update steps via minibatch sampling, and finally flush the buffers.

\begin{algorithm}[H]
\caption{Pseudocode of DACMP training procedure (Part I)}
\label{alg}
\SetAlgoLined % 这里建议显式声明
\footnotesize
\DontPrintSemicolon
% --- 关键：局部重定义 While，把最后一个参数 end 设为空 {} ---
\SetKwProg{While}{while}{ \textbf{do}}{} 

% 初始化部分
\KwIn{Prior policy $\pi_\text{prior}$ (composed of RRT* and PID)}
\KwOut{Optimized policies $\pi_{\text{m}}^*, \pi_{\text{b}}^*$}
\BlankLine
Initialize dual-agent actor parameters $\theta_{\text{m}}, \theta_{\text{b}}$ and critic parameters $\phi_\text{m}, \phi_\text{b}$\;
Initialize experience replay buffers $\mathcal{D}_j$ for $j \in \{\text{m}, \text{b}\}$\;
Initialize update counter $k=0$, episode counter $m=0$\;

\BlankLine
\While{\textnormal{episode} $m \leq M$}
{
  $m \leftarrow m + 1$\;
  \tcc{Phase 1: Data Collection and Storage with TESG}
  Reset environment to initial state $\boldsymbol{s}_0$\;
  
  % \tcc{Interaction Phase}
  \For{\textnormal{step} $t=0$ \KwTo $T-1$}{
    % \tcc{Phase 1: TESG Action Selection}
    Obtain DRL actions: $\{a^\text{m}_t, a_t^\text{b}\}_\text{DRL}$, where $a_t^\text{m} \sim \pi_{\theta_\text{m}}(s_t^\text{m})$ and $a_t^\text{b} \sim \pi_{\theta_\text{b}}(s_t^\text{b})$\;
    Obtain prior actions: $\{a^\text{m}_t, a_t^\text{b}\}_\text{prior} \leftarrow \pi_\text{prior}(\boldsymbol{s_t})$\;
    Sample random variable $\zeta \sim \mathcal{U}(0, 1)$\;
    Calculate switching threshold $p(k)$ according to Eq. \eqref{eq24}\;
    \eIf{$\zeta \leq p(k)$}{
      $\boldsymbol{a}_t \leftarrow \{a^\text{m}_t, a_t^\text{b}\}_\text{DRL}$ %\tcc*[r]{Exploit DRL}
    }{
      $\boldsymbol{a}_t \leftarrow \{a^\text{m}_t, a_t^\text{b}\}_\text{prior}$ %\tcc*[r]{Follow Prior}
    }

    % \tcc{Phase 2: Execution}
    Execute $\boldsymbol{a}_t$, observe $\boldsymbol{s}_{t+1}$ and calculate $\boldsymbol{r}_t$\;
    Store transition $\langle s^j_t, a^j_t, \log\pi^j_t, s^j_{t+1}, r^j_t \rangle$ into $\mathcal{D}_j$ for $j \in \{\text{m}, \text{b}\}$\;    $s_t \leftarrow s_{t+1}$\;
  }
}
\end{algorithm}

\addtocounter{algocf}{-1} 
\renewcommand{\thealgocf}{\arabic{algocf}} % 恢复正常编号
\begin{algorithm}[H]
\caption{Pseudocode of DACMP training procedure (Part II)}
\SetAlgoLined 
\SetAlgoLongEnd 
\footnotesize
\DontPrintSemicolon
\setcounter{AlgoLine}{20} % 接续 Part I 的行号
% 我们使用一个小技巧：定义一个不带标题的 Dummy 块，或者直接写逻辑延续
\SetKwBlock{Dummy}{}{}
\def\nl{\relax} 
\Dummy{
  \vspace{-1.2em} % 把 Dummy 产生的空行顶回去
  \let\nl\oldnl % 恢复行号（需在导言区备份 \let\oldnl\nl）
  \If{\textnormal{size}($\mathcal{D}_j$) $\geq$ \textnormal{Experience buffer size} $C$ }{
    $k \leftarrow k + 1$\;
    \tcc{Phase 2: Off-Policy Dual-Agent Update}
    \For{\textnormal{agent} $j \in \{\textnormal{m}, \textnormal{b}\}$}{
      \For{\textnormal{iteration} $i=1$ \KwTo $K_\textnormal{update}$}{
        Sample random minibatch of size $N$ from $\mathcal{D}_j$\;
        Update Critic $\phi_j$ by minimizing value function loss\;
        Update Actor $\theta_j$ by maximizing clipped objective in Eq. \eqref{eq7}\;
      }
    }
    \tcc{Phase 3: Experience Buffers Flush}
    Clear $\mathcal{D}_j$ for $j \in \{\text{m}, \text{b}\}$\;
  }
}
\let\nl\oldnl % 恢复行号（需在导言区备份 \let\oldnl\nl）
\textbf{end}
\end{algorithm}
\renewcommand{\thealgocf}{\arabic{algocf}} % 恢复正常编号

\section{Hyperparameters of the reward functions}
\label{appC}
In this appendix, we summarize the hyperparameters used in the reward functions for both the manipulator and the base spacecraft, including weighting coefficients, thresholds, and tolerances. The detailed values are listed in Table \ref{tableap2}.

\begin{table}[!htb]
  \caption{Hyperparameters of the reward functions.}
  \label{tableap2}
  \footnotesize
  \centering
  \begin{tabular}{l l l l}
  %% Tabular cells are separated by &
    \toprule
    Agent & Symbol & Value & Description \\     % Table header
    \midrule
    \multirow{8}{*}{Space Manipulator}
    & $k_{\text{pos}}^\text{m}$ & $0.5$ & Position error weight \\
    & $k_{\text{ori}}^\text{m}$ & $0.125$ & Orientation error weight \\
    & $k_{\text{smth}}^\text{m}$ & $0.1$ & Smoothness penalty coefficient \\
    & $\delta_{\dot{q}}$ & $2.0$ rad/s & Joint velocity smoothness tolerance \\
    & $k_{\text{aln}}^\text{m}$ & $0.15$ & Orientation alignment reward coefficient \\
    & $k_{\text{done}}^\text{m}$ & $0.1$ & Completion reward coefficient \\
    & $\varepsilon_\text{pos}$ & $0.05$ m & Position threshold \\
    & $\varepsilon_\text{ori}$ & $0.1$ rad & Orientation threshold \\
    % \hspace*{\fill} \\
    \addlinespace
    % \midrule
    \multirow{4}{*}{Base Spacecraft}
    & $k_{\text{att}}^\text{b}$ & $2.5$ & Base attitude penalty coefficient \\
    & $k_{\text{var}}^\text{b}$ & $2.5$ & Base attitude variation reward coefficient \\
    & $k_{\text{done}}^\text{b}$ & $0.2$ & Base completion reward coefficient \\
    & $\varepsilon_\text{att}$ & $0.05$ rad & Attitude threshold \\
    \bottomrule
  \end{tabular}
\end{table}

\section*{Acknowledgement}
This work was kindly supported by the Fundamental and Interdisciplinary Disciplines Breakthrough Plan of the Ministry of Education of China through grant No.~JYB2025XDXM207, National Natural Science Foundation of China through grant No.~62403162 and Joint Funds of the National Natural Science Foundation of China through grant No.~U23A20346.

\end{document}